%% file: main.tex
\documentclass[letterpaper]{article} 
\usepackage{aaai2027}  
\usepackage[hyphens]{url}  
\usepackage{graphicx} 
\usepackage{natbib}  
\usepackage{caption} 
\usepackage{algorithm}
\usepackage{algorithmic}

\usepackage{newfloat}
\usepackage{listings}
\DeclareCaptionStyle{ruled}{labelfont=normalfont,labelsep=colon,strut=off} 
\floatstyle{ruled}
\newfloat{listing}{tb}{lst}{}
\floatname{listing}{Listing}

\usepackage{booktabs}
\usepackage{amsmath,amssymb}
\nocopyright

\title{Don't Waste the Noise:\\ Importance-Guided Perturbation Allocation under Joint Global and Local Constraints
}
\author{
    Melika Shirian\textsuperscript{\rm 1}\equalcontrib,
    Kianoosh Vadaei\textsuperscript{\rm 1}\equalcontrib
}

\affiliations{
    \textsuperscript{\rm 1}Independent Researcher\\
    melika.shi@mehr.ui.ac.ir, k.vadaei@mehr.ui.ac.ir
}

\begin{document}

\maketitle

\begin{abstract}
Adversarial optimization under a shared $\ell_1$ budget requires deciding not only how much perturbation to use, but also where that limited budget should be spent. This allocation problem becomes particularly important when individual input coordinates are subject to local magnitude constraints, which restrict the extent to which perturbation can be concentrated on a small number of locations. We introduce an importance-guided allocation mechanism that uses a fixed clean-gradient prior to steer perturbation toward model-sensitive regions while leaving the feasible perturbation set unchanged. A centered allocation objective encourages perturbation at above-average importance locations and discourages unnecessary expenditure elsewhere, thereby redistributing rather than enlarging the available budget. Across ten robust model--dataset configurations under a common capacity-limited threat setting, the proposed method improves attack success over matched APGD- and PMA-based baselines by $2.52$ to $17.70$ percentage points. Allocation analysis shows that these gains are accompanied by substantially greater perturbation mass in high-importance regions without increased global $\ell_1$ consumption. Mechanism ablations further show that centered non-uniform redistribution provides part of the benefit, while model-derived importance yields an additional improvement. These results identify perturbation allocation as a distinct and practically relevant dimension of adversarial optimization under shared-budget, locally constrained threat models.
\end{abstract}

\begin{figure}[t]
    \centering
    \includegraphics[width=\columnwidth]{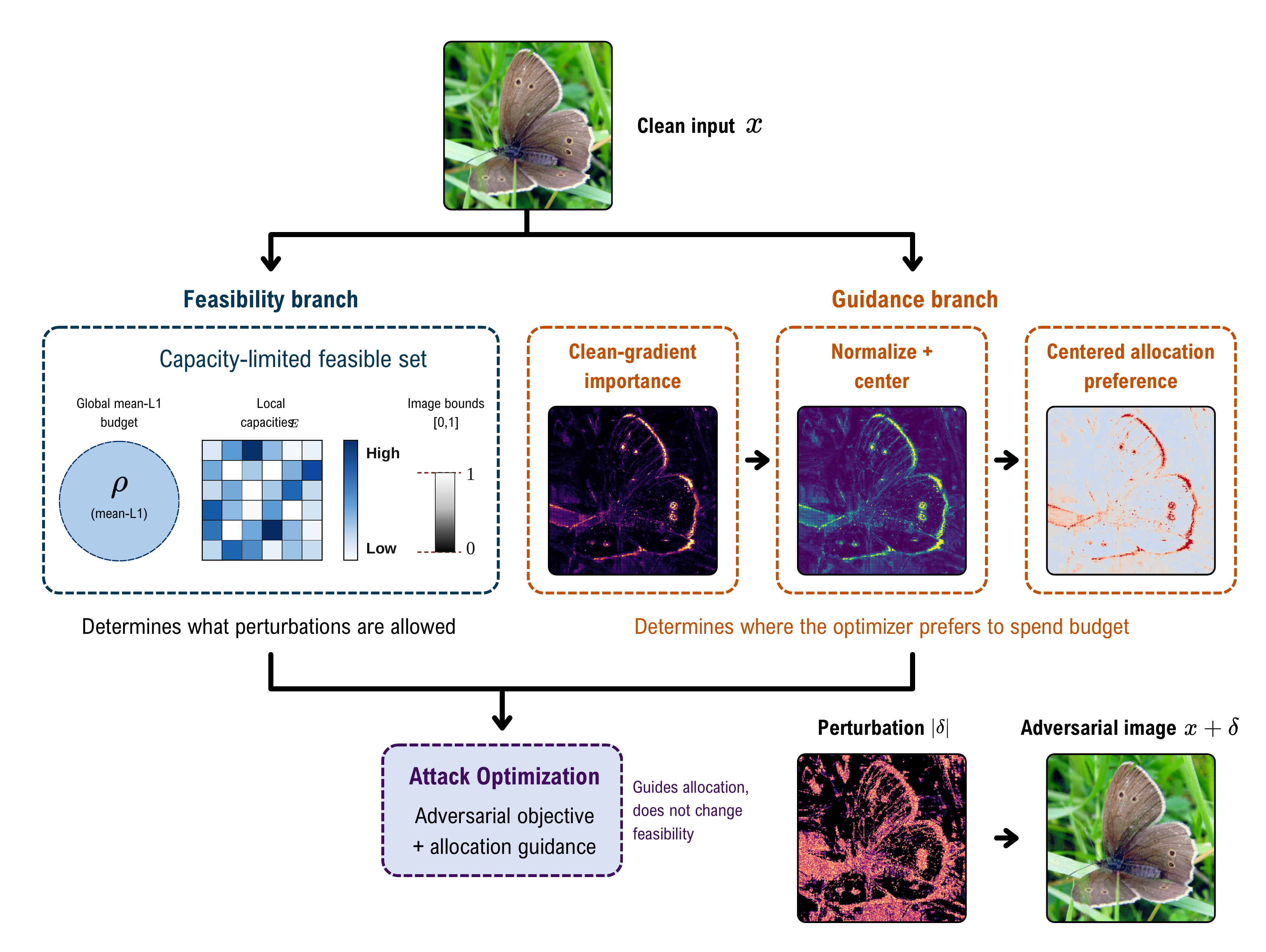}
    \caption{Overview of importance-guided perturbation budget allocation.
    The capacity-limited threat model determines which perturbations are
    feasible through a shared mean-\(L_1\) budget, coordinate-wise capacities,
    and image-range bounds. Independently, a clean-gradient importance prior
    is normalized and centered to guide where the perturbation
    budget is spent. The allocation signal changes the optimization preference
    without changing the feasible perturbation set.}
    \label{fig:method_overview}
\end{figure}

\input{Sections/introduction}


\input{Sections/related_work}
\input{Sections/Importance_Aware_Perturbation_Budget_Allocation}

\input{Sections/experimental_setup}

\input{Sections/results}

\input{Sections/limitations}

\input{Sections/conclusion}

\bibliography{references}

\clearpage

\appendix
\setcounter{secnumdepth}{2}

\input{Appendix/full_threat_model_details}
\input{Appendix/projection_and_optimization_details}
\input{Appendix/capacity_map_construction}
\input{Appendix/baseline_adaptation_details}
\input{Appendix/dataset_model_evaluation_populations}
\input{Appendix/complete_hyperparameters}
\input{Appendix/lambda_selection_and_sensitivity}
\input{Appendix/full_mechanism_ablations}
\input{Appendix/allocation_diagnostics}
\input{Appendix/perceptual_and_norm_diagnostics}
\input{Appendix/computational_cost}
\input{Appendix/statistical_analysis}
\input{Appendix/descriptive_cross_regime_results}
\input{Appendix/constraint_validity_checks}
\input{Appendix/additional_qualitative_examples}

\end{document}

%% file: Sections/introduction.tex
\section{Introduction}
\label{sec:introduction}

Adversarial attacks under $\ell_1$-bounded threat models have become
increasingly effective at identifying vulnerabilities in image classifiers.
Unlike an $\ell_\infty$ constraint, which independently limits every input
coordinate, an $\ell_1$ constraint provides a shared perturbation budget
that may be distributed unevenly across the input
\cite{croce2023mindboxl1apgdsparse,pintor2021fastminimumnormadversarialattacks}.
Specialized methods such as $\ell_1$-APGD exploit this geometry by
concentrating perturbation on influential coordinates
\cite{croce2023mindboxl1apgdsparse}, while sparse attacks similarly exploit
small subsets of sensitive input components
\cite{modas2019sparsefoolpixelsmakebig,croce2019sparseimperceivableadversarialattacks}.

This flexibility becomes more limited when individual coordinates are also
subject to local magnitude constraints. Once attractive coordinates approach
their local capacities, useful perturbation must be redirected elsewhere.
Adversarial optimization therefore becomes not only a problem of finding a
useful update direction, but also one of deciding \emph{where the available
budget should be spent}.

This motivates our central question:

\begin{quote}
\emph{When a shared $\ell_1$ perturbation budget must be distributed among
locally capacity-limited coordinates, can explicit guidance about where that
budget is spent improve adversarial optimization?}
\end{quote}

We study this question by separating two roles: \emph{feasibility}
determines which perturbations are allowed, while \emph{allocation
preference} determines where an optimizer prefers to spend the available
budget. Figure~\ref{fig:method_overview} summarizes this distinction.

\subsection{Capacity-Limited $\ell_1$ Optimization}

We consider a shared global mean-$\ell_1$ budget together with
coordinate-wise magnitude capacities and valid image-range constraints.
The global constraint limits the total available perturbation, while the
local capacities limit how much of that budget may be concentrated at any
individual coordinate
\cite{croce2019sparseimperceivableadversarialattacks}. When a useful
coordinate reaches its local capacity, the optimizer must distribute the
remaining budget across other feasible coordinates.

This creates a budget-allocation problem that is related to, but distinct
from, sparse optimization. Sparse methods primarily determine which
coordinates should receive large changes, whereas capacity-limited
optimization must also determine how a shared budget should be distributed
when individual coordinates cannot absorb arbitrarily large perturbations.
The formal feasible set is defined in the Method section.

All methods in our main evaluation operate under the same capacity-limited
feasible set. Separately parameterized conventional and capacity-limited
$\ell_1$ results are retained as descriptive comparisons in
Appendix~\ref{app:cross_regime}; because their global budget
parameterizations differ, they are not used to isolate the causal effect of
local capacities.

\subsection{Importance-Guided Budget Allocation}

We address the allocation problem using a fixed, model-derived importance
prior. Given a clean input, we construct a spatial importance map from the
model gradient \cite{simonyan2014deepinsideconvolutionalnetworks}, normalize
it, and center it around its spatial mean. The resulting allocation reward
is added to the adversarial objective to encourage perturbation magnitude to
shift toward above-average importance locations.

Crucially, the importance prior does not change the threat model. It does
not mask low-importance coordinates, enlarge local capacities, or provide
additional perturbation budget. Instead, it changes only the optimization
preference within the same feasible region. This distinguishes our approach
from attacks that explicitly select a sparse support or restrict
perturbations to salient regions
\cite{papernot2015limitationsdeeplearningadversarial,Dong_2020_CVPR}.

Centering is important because it prevents the allocation term from simply
rewarding larger perturbations everywhere. Instead, it creates a relative
preference that encourages perturbation mass to move from below-average
toward above-average importance locations. The full importance construction,
allocation objective, and optimization procedure are defined in the Method
section.

\subsection{Main Findings and Contributions}

We evaluate the proposed mechanism across ten robust model--dataset
configurations spanning CIFAR-10, CIFAR-100, and ImageNet. Under the common
capacity-limited threat model, our method improves attack success over the
highest-ASR matched APGD/PMA baseline in every configuration, with gains of
$2.52$--$17.70$ percentage points.

These improvements are accompanied by a systematic change in budget
allocation. Matched baselines place approximately $10.6$--$11.3\%$ of
their perturbation mass in the highest-importance $10\%$ of spatial
locations, whereas our method places $19.2$--$30.4\%$ there while remaining
within the same global mean-$\ell_1$ budget. Mechanism ablations further
separate two effects: centered non-uniform redistribution itself improves
optimization, while alignment with model-derived importance provides an
additional gain.

Our main contributions are:

\begin{itemize}
    \item \textbf{Perturbation allocation as an optimization problem.}
    We study how a shared global mean-$\ell_1$ budget should be distributed
    when individual coordinates are subject to local magnitude capacities,
    explicitly separating feasibility from allocation preference.

    \item \textbf{Centered importance-guided allocation.}
    We introduce a soft, model-derived allocation objective that guides
    perturbation toward more informative regions without masking coordinates
    or changing the feasible perturbation set.

    \item \textbf{Mechanistic empirical evidence.}
    Across ten robust model--dataset configurations, the mechanism improves
    matched APGD/PMA-style baselines. Allocation diagnostics and controlled
    ablations further distinguish the contribution of non-uniform
    redistribution from the additional benefit of meaningful importance
    alignment.
\end{itemize}

We do not claim universal superiority over specialized $\ell_1$ attacks.
Adapted $\ell_1$-APGD remains stronger in the separate reference comparisons
reported in Appendix~\ref{app:cross_regime}. Rather, our results identify
\emph{perturbation allocation} as a distinct optimization dimension that can
substantially improve matched adversarial optimizers under the
capacity-limited setting studied here.

%% file: Sections/related_work.tex
\section{Related Work}
\label{sec:related_work}

\paragraph{Gradient-based adversarial optimization.}
Gradient-based attacks have advanced through improvements in optimization
procedures, attack objectives, and evaluation protocols
\cite{croce2021robustbenchstandardizedadversarialrobustness,Cin__2025}.
Projected gradient descent (PGD) became a standard approach for
norm-constrained adversarial optimization
\cite{madry2019deeplearningmodelsresistant}, while methods such as C\&W
\cite{carlini2017evaluatingrobustnessneuralnetworks} and DeepFool
\cite{moosavidezfooli2016deepfoolsimpleaccuratemethod} explored alternative
objective and boundary-based formulations. Auto-PGD (APGD) further improves
iterative optimization through adaptive step-size control and related
stabilization mechanisms, and forms a core component of AutoAttack
\cite{croce2020reliable}. Probability Margin Attack (PMA) shows that attack
strength can also depend strongly on the adversarial objective itself,
using a probability-space margin rather than a logit-space formulation
\cite{xie2025million}. Our work complements these directions by studying
how a shared perturbation budget is allocated across the input under local
capacity constraints.

\paragraph{$\ell_1$ and sparse attacks.}
Sparse and $\ell_1$-oriented attacks explicitly exploit the geometry of
perturbation concentration \cite{croce2022sparseRS}. EAD incorporates an
$\ell_1$ penalty through an elastic-net formulation
\cite{chen2018ead}, while SparseFool uses local decision-boundary geometry
to construct perturbations affecting relatively few input components
\cite{modas2019sparsefool}. Croce and Hein studied sparse attacks with
additional component-wise magnitude constraints
\cite{croce2019sparseimperceivableadversarialattacks}, and later introduced $\ell_1$-APGD, which
adaptively controls update sparsity to better exploit $\ell_1$ geometry
\cite{croce2019sparseimperceivableadversarialattacks}. Fast Minimum-Norm (FMN) attacks similarly support
$\ell_1$ optimization by adapting the norm constraint while searching for
small adversarial perturbations \cite{pintor2021fastminimumnormadversarialattacks}.

Our setting is closely related because it combines a shared $\ell_1$-type
budget with coordinate-wise magnitude limits. We do not treat these local
capacity constraints themselves as the novelty of the work. Instead, we
study the optimization problem that arises \emph{within} such a feasible
set: how the shared budget should be distributed when individual
coordinates cannot absorb arbitrarily large perturbations.

\paragraph{Importance-guided and structured perturbations.}
Prior attacks have used saliency or spatial structure to influence where
perturbations are applied. JSMA constructs a saliency map to identify
influential input features \cite{papernot2016limitations}; GreedyFool
combines gradient information with a distortion map to select candidate
pixels for sparse modification \cite{dong2020greedyfool}; and
superpixel-guided or saliency-based attacks restrict or prioritize
perturbations in informative spatial regions
\cite{dong2020superpixel,dai2023saliency}. More generally, gradient-based
saliency provides a simple estimate of local model sensitivity
\cite{simonyan2014deepinsideconvolutionalnetworks}. Structured adversarial attacks additionally impose
spatial, group-sparse, or related regularities on the perturbation
\cite{xu2019structured}.

Our mechanism differs from hard support selection or importance-dependent
constraints. The importance map does not determine which coordinates are
feasible and does not increase the admissible magnitude of
high-importance locations. Instead, it enters the attack objective as a
soft, centered preference that encourages perturbation mass to be
redistributed toward more important regions while retaining access to the
same feasible set.

This positioning makes the contribution narrower than introducing a new
saliency attack or a new capacity constraint: we study \emph{perturbation
allocation} as a distinct optimization component under a shared-budget,
locally constrained threat model.

%% file: Sections/Importance_Aware_Perturbation_Budget_Allocation.tex
\section{Importance-Guided Perturbation Budget Allocation}
\label{sec:method}

We consider adversarial optimization under a shared global perturbation
budget and coordinate-wise magnitude constraints. The feasible set specifies
which perturbations are admissible, while our allocation objective provides
a soft preference over where the available budget should be spent.
Model-derived importance affects only this optimization preference and does
not modify the admissible perturbation region.

\subsection{Threat Model and Local Capacities}
\label{sec:threat_model}

Let $x \in [0,1]^D$ be a correctly classified input with ground-truth label
$y$, where $D$ denotes the number of scalar input coordinates. We seek an
adversarial perturbation $\delta$ satisfying
\begin{equation}
\mathcal{C}(x,E,b)
=
\left\{
\delta :
x+\delta \in [0,1]^D,\;
|\delta_i| \leq E_i,\;
\frac{\|\delta\|_1}{D} \leq b
\right\},
\label{eq:feasible_set}
\end{equation}
where $b$ is a global mean-$\ell_1$ budget and $E_i$ is the local magnitude
capacity of coordinate $i$.

The global constraint couples all coordinates through a shared perturbation
budget, while the local capacities limit how much of that budget may be
absorbed by any individual coordinate. Thus, once locally attractive
coordinates approach their capacities, the optimizer must distribute useful
perturbation over other feasible coordinates.

\paragraph{Heterogeneous local capacities.}
In the main experiments, capacities are generated spatially as
\begin{equation}
E_p
=
\mu_E
+
\sigma_E
\Phi^{-1}(U_p),
\qquad
U_p \sim \mathcal{U}(u_{\min},u_{\max}),
\label{eq:capacity_sampling}
\end{equation}
where $p$ indexes spatial locations and $\Phi^{-1}$ is the inverse CDF of
the standard normal distribution. We use
$u_{\min}=10^{-3}$ and $u_{\max}=1-10^{-3}$ to avoid extreme tails, and
share the sampled capacity across the RGB channels of each spatial
location.

The capacity map is generated independently of model-derived importance:
$E$ specifies how much perturbation is feasible at a location, not whether
that location should be preferred by the optimizer. All matched methods use
the same realization of $E$ and the same global budget $b$. Further
construction details are provided in
Appendix~\ref{app:capacity_details}.

\subsection{Clean-Gradient Importance Prior}
\label{sec:importance_prior}

We construct a fixed spatial importance prior from the clean input. Let
$z_y(x)$ denote the logit of the ground-truth class. For spatial location
$p$ with $C$ input channels, we define
\begin{equation}
s_p
=
\frac{1}{C}
\sum_{c=1}^{C}
\left|
\frac{\partial z_y(x)}
     {\partial x_{c,p}}
\right|.
\label{eq:raw_importance}
\end{equation}

These scores provide a local estimate of model sensitivity
\cite{simonyan2014deepinsideconvolutionalnetworks}. To reduce the influence
of isolated gradient spikes, we clip them at an upper quantile $q$:
\begin{equation}
\hat{s}_p
=
\min\left\{
s_p,\,
Q_q(s)
\right\},
\label{eq:importance_clip}
\end{equation}
where $Q_q(s)$ is the $q$-th quantile of the spatial scores.

We then normalize by the spatial mean,
\begin{equation}
u_p
=
\frac{\hat{s}_p}
     {\operatorname{mean}(\hat{s})+\epsilon},
\label{eq:importance_normalize}
\end{equation}
and construct a bounded importance map
\begin{equation}
M_p
=
\operatorname{clip}
\left(
a u_p^{\gamma},
M_{\min},
M_{\max}
\right),
\label{eq:importance_map}
\end{equation}
where $\gamma$ controls the sharpness of the importance distribution and
$a>0$ is chosen so that
\begin{equation}
\operatorname{mean}(M)=1.
\label{eq:importance_mean}
\end{equation}

The resulting map is shared across channels at each spatial location,
computed once from the clean input, and held fixed throughout the attack.
It is therefore an allocation prior rather than a dynamically updated
attack trajectory. Complete importance-map hyperparameters are reported in
Appendix~\ref{app:hyperparameters}.

\subsection{Centered Allocation Objective}
\label{sec:allocation_objective}

We incorporate the importance prior through a soft allocation reward. The
attack objective is
\begin{equation}
J_{\mathrm{ours}}(\delta)
=
J_{\mathrm{adv}}(x+\delta,y)
+
\lambda R_{\mathrm{alloc}}(\delta,M),
\label{eq:ours_objective}
\end{equation}
where $J_{\mathrm{adv}}$ is the adversarial classification objective and
$\lambda$ controls the strength of allocation guidance.

Because $M$ is normalized to have unit spatial mean, $M_i-1$ distinguishes
above-average from below-average importance. We define
\begin{equation}
R_{\mathrm{alloc}}(\delta,M)
=
\frac{1}{Db}
\sum_{i=1}^{D}
(M_i-1)
\sqrt{\delta_i^2+\tau^2},
\label{eq:allocation_reward}
\end{equation}
where $\sqrt{\delta_i^2+\tau^2}$ is a smooth approximation to
$|\delta_i|$ and $\tau>0$ is a small smoothing constant.

For $M_i>1$, increasing perturbation magnitude contributes positively to
the reward; for $M_i<1$, it is penalized. The normalization by $Db$ scales
the allocation term relative to the available global perturbation budget.

Centering is essential to the intended mechanism. Using $M_i$ directly
would assign a positive magnitude reward at every coordinate, whereas
$M_i-1$ creates a relative preference that encourages perturbation mass to
shift from below-average toward above-average importance locations.

Importantly, $R_{\mathrm{alloc}}$ does not alter
$\mathcal{C}(x,E,b)$. Low-importance coordinates remain feasible,
high-importance coordinates receive no additional capacity, and no hard
support is imposed. The mechanism therefore changes how the optimizer
navigates the feasible set rather than changing the set itself.

\subsection{Attack Optimization}
\label{sec:attack_optimization}

Starting from a feasible initialization $\delta_0$, we directly optimize
Equation~\ref{eq:ours_objective}. At iteration $t$, we compute
\begin{equation}
g_t
=
\nabla_{\delta}
J_{\mathrm{ours}}(\delta_t),
\label{eq:joint_gradient}
\end{equation}
normalize the gradient by its mean absolute magnitude, and accumulate
momentum following momentum-based adversarial optimization
\cite{dong2018momentum}. The perturbation is updated in the sign direction
of the accumulated gradient, with a cosine-decayed update magnitude
\cite{loshchilov2017sgdr}.

After each update, feasibility is restored using the same constraint
operator employed for the matched attacks. We sequentially
\begin{enumerate}
    \item clip the perturbation to the coordinate-wise capacities $E$;
    \item enforce the valid image range $x+\delta\in[0,1]^D$; and
    \item project onto the global $\ell_1$ ball of radius $Db$.
\end{enumerate}
The final step uses the standard Euclidean projection onto an $\ell_1$ ball
\cite{duchi2008efficient}. Initialization, step-size, momentum, restart, and
projection details are provided in
Appendix~\ref{app:optimization_details}. The corresponding exact-projection
control is reported in Appendix~\ref{app:projection_control}.

\subsection{Candidate Selection}
\label{sec:candidate_selection}

We retain the strongest feasible candidate encountered during optimization
rather than returning only the final iterate. Candidates are ranked by the
probability margin \cite{xie2025million},
\begin{equation}
m_{\mathrm{prob}}(x+\delta)
=
\max_{k\neq y}
p_k(x+\delta)
-
p_y(x+\delta),
\label{eq:probability_margin}
\end{equation}
where $p_k(x+\delta)$ is the predicted probability of class $k$.

The feasible candidate with the largest margin is returned. This criterion
is used only for candidate selection and does not replace the optimization
objective in Equation~\ref{eq:ours_objective}. Unless explicitly varied in
the objective ablations, the attack trajectory is determined by
$J_{\mathrm{adv}}$ together with the centered allocation reward.

The resulting construction keeps the two quantities central to our analysis
separate: $E$ determines local feasibility, while $M$ supplies an
independent preference over budget allocation. The mechanism ablations in
the Results and Appendix~\ref{app:mechanism_ablations} test the roles of
centering, non-uniform redistribution, and importance quality separately.

%% file: Sections/experimental_setup.tex
\begin{figure*}[t]
    \centering
    \includegraphics[
        width=0.78\textwidth,
        height=0.19\textheight,
        keepaspectratio
    ]{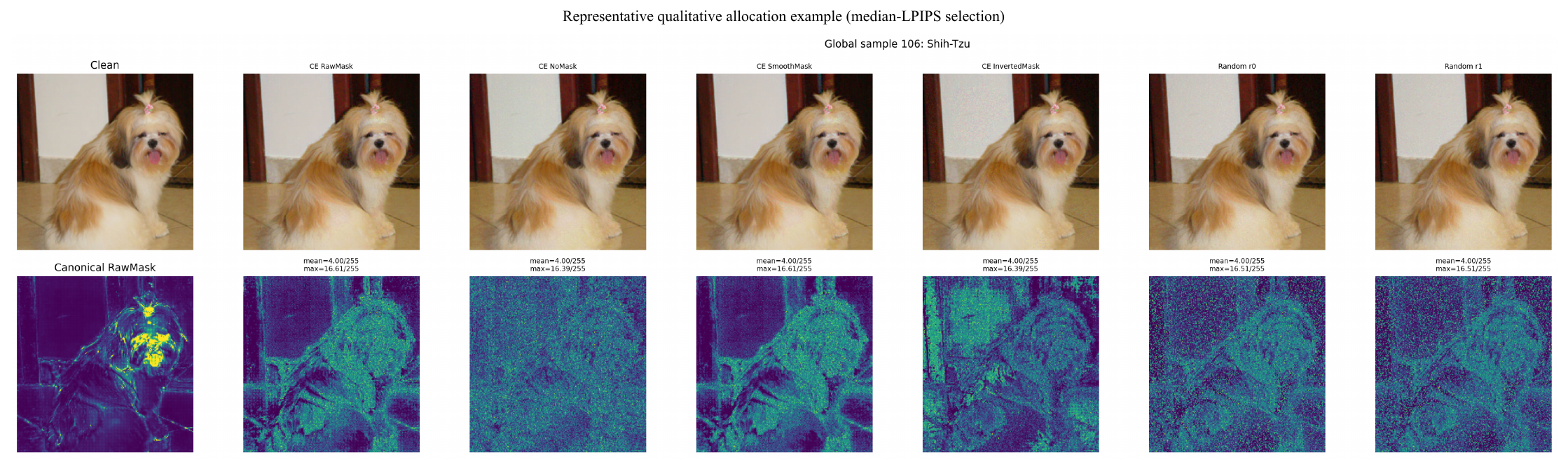}
    \caption{Qualitative example of importance-aware perturbation allocation.
    The clean-gradient importance map provides a fixed spatial prior, while
    the perturbation-magnitude maps show how different mechanism variants
    distribute perturbation under identical constraints. Ours exhibits
    stronger alignment with regions emphasized by the importance prior,
    whereas removing, randomizing, or reversing the guidance changes the
    resulting allocation pattern.}
    \label{fig:qualitative_allocation}
\end{figure*}

\section{Experimental Setup}
\label{sec:experimental_setup}

Our main evaluation compares attacks \emph{within a common
capacity-limited feasible set}. For each model and evaluation sample, all
matched methods use the same global perturbation budget, local-capacity
realization, image-range constraints, and evaluation population. A separate
comparison between conventional and capacity-limited $\ell_1$ settings is
reported descriptively in Appendix~\ref{app:cross_regime}; because those
settings use different global budget parameterizations, it is not used to
isolate the effect of local capacities.

\subsection{Datasets and Models}
\label{sec:datasets_models}

We evaluate on CIFAR-10, CIFAR-100
\cite{krizhevsky2009learning}, and ImageNet
\cite{deng2009imagenet}, using ten robust model--dataset configurations
spanning convolutional and transformer-based architectures.

On CIFAR-10, we evaluate WRN-94-16 from Bartoldson et al.~
\cite{bartoldson2024limits}, WRN-28-10 from Cui et al.~
\cite{cui2024decoupled}, and XCiT-S12 from Debenedetti et al.~
\cite{debenedetti2023light}. On CIFAR-100, we evaluate WRN-28-10
\cite{cui2024decoupled}, XCiT-S12
\cite{debenedetti2023light}, and WRN-70-16 from Wang et al.~
\cite{wang2023better}. On ImageNet, we evaluate ConvNeXt-L
\cite{liu2025comprehensive}, the robust model of Peng et al.~
\cite{peng2023robust}, ViT-B with a convolutional stem
\cite{singh2023revisiting}, and Swin-L from MIMIR
\cite{xu2026mimir}.

Table~\ref{tab:evaluation_panel} reports the attacked population for each
configuration. All methods compared within a row operate on exactly the
same inputs and labels.

\begin{table}[t]
\centering
\caption{\textbf{Evaluation panel.} $N$ is the number of clean-correct
inputs attacked in each configuration; all methods within a row use the
same evaluation population.}
\label{tab:evaluation_panel}
\small
\begin{tabular}{llr}
\toprule
Dataset & Model & $N$ \\
\midrule
CIFAR-10  & WRN-94-16      & 9,368 \\
CIFAR-10  & WRN-28-10      & 7,500 \\
CIFAR-10  & XCiT-S12       & 9,006 \\
\midrule
CIFAR-100 & WRN-28-10      & 7,383 \\
CIFAR-100 & XCiT-S12       & 6,734 \\
CIFAR-100 & WRN-70-16      & 7,522 \\
\midrule
ImageNet  & ConvNeXt-L     & 1,000 \\
ImageNet  & Peng et al.    & 1,000 \\
ImageNet  & ViT-B-ConvStem & 1,000 \\
ImageNet  & Swin-L         & 1,000 \\
\bottomrule
\end{tabular}
\end{table}

For ImageNet, we use 1,000 clean-correct validation images selected in a
fixed deterministic order for each model. Complete checkpoint identifiers,
preprocessing details, clean-correct population sizes, and sampling
information are provided in
Appendix~\ref{app:evaluation_populations}.

\subsection{Capacity-Limited Threat Setting}
\label{sec:experimental_threat}

All main comparisons use the feasible set defined in
Equation~\ref{eq:feasible_set}. We set the global mean-$\ell_1$ budget to
$b=4/255$. The heterogeneous capacity map follows
Equation~\ref{eq:capacity_sampling} with
$\mu_E=12/255$ and $\sigma_E=2/255$, with each spatial capacity shared
across RGB channels.

For a given sample, all matched methods receive the same capacity
realization and enforce the same global mean-$\ell_1$, coordinate-wise,
and image-range constraints. Where direct initialization matching is
applicable, they also use the same initial perturbation.

The capacity map is generated independently of the importance prior, so
model-derived importance affects optimization preference rather than local
feasibility. Capacity construction and constraint-validity checks are
reported in Appendix~\ref{app:capacity_details} and
Appendix~\ref{app:constraint_validity}, respectively.

\subsection{Matched Baselines}
\label{sec:baselines}

We compare against three gradient-based baselines adapted to the same
capacity-limited feasible set:

\begin{itemize}
    \item \textbf{APGD-CE}, using cross-entropy;
    \item \textbf{APGD-DLR}, using the difference-of-logits-ratio objective
    \cite{croce2020reliable}; and
    \item \textbf{PMA}, using the probability-margin objective
    \cite{xie2025million}.
\end{itemize}

The adaptations enforce the common constraints while retaining each
baseline's characteristic optimization procedure or objective. Our
reference attack uses untargeted cross-entropy augmented with the centered
allocation reward in Equation~\ref{eq:allocation_reward}.

Unless otherwise stated, all attacks use 50 optimization steps and one
restart. Full adaptation details and attack hyperparameters are provided in
Appendix~\ref{app:baseline_adaptations} and
Appendix~\ref{app:hyperparameters}.

These baselines are used specifically to test allocation guidance under a
matched feasible set. Results for specialized $\ell_1$-APGD and
FMN-$\ell_1$ are reported separately in
Appendix~\ref{app:cross_regime} and are not part of the universal baseline
comparison.

\subsection{Allocation-Weight Selection}
\label{sec:lambda_selection}

We select the allocation strength $\lambda$ independently for each dataset
using 100 held-out clean-correct samples. The candidate set is
$\{0.10,0.20,0.35,0.50,0.75\}$, with ASR used as the selection criterion.
This yields
$\lambda=0.5$ for CIFAR-10,
$\lambda=0.5$ for CIFAR-100, and
$\lambda=0.2$ for ImageNet.

The selected value is then frozen across models and reported evaluations
within that dataset. Complete selection and sensitivity results are given in
Appendix~\ref{app:lambda_selection}.

\subsection{Evaluation Metrics}
\label{sec:evaluation_metrics}

Our primary effectiveness metric is conditional attack success rate (ASR),
computed over inputs correctly classified by the corresponding clean model.

To measure perturbation allocation, let $\mathcal{S}_{\alpha}(M)$ denote
the top $\alpha$ fraction of spatial locations ranked by the clean-gradient
importance map. We define
\begin{equation}
\operatorname{Mass}_{\alpha}(\delta,M)
=
\frac{
\sum_{i\in\mathcal{S}_{\alpha}(M)} |\delta_i|
}{
\sum_{i=1}^{D} |\delta_i|
},
\label{eq:importance_mass_metric}
\end{equation}
where the spatial ranking is shared across the corresponding RGB
coordinates. Our primary allocation statistic uses $\alpha=0.10$.

We additionally report cumulative importance-ranked perturbation mass and
the realized global magnitude
$255\|\delta\|_1/D$ to verify that differences in attack effectiveness are
not explained by greater consumption of the global $\ell_1$ budget.
Complementary allocation statistics are reported in
Appendix~\ref{app:allocation_diagnostics}, while norm and perceptual
diagnostics are reported in
Appendix~\ref{app:perceptual_diagnostics}.

\subsection{Paired Statistical Analysis}
\label{sec:statistical_analysis}

Because methods within each configuration attack the same clean-correct
inputs, performance comparisons are paired. We report paired ASR
differences with 95\% confidence intervals and evaluate binary success
differences using McNemar's test \cite{mcnemar1947note}. Multiple
comparisons are corrected using Holm's procedure
\cite{holm1979simple}.

The main text emphasizes effect sizes and confidence intervals; complete
pairwise statistics and corrected $p$-values are reported in
Appendix~\ref{app:statistics}.

\subsection{Computational Accounting and Reproducibility}
\label{sec:compute_protocol}

We measure computational cost using wall-clock runtime and forward/backward
model evaluations. Constructing the clean-gradient importance prior requires
one additional backward evaluation before iterative optimization. A
compute-matched control tests whether this additional evaluation explains
the observed gain.

Full timing and evaluation counts are reported in
Appendix~\ref{app:compute}. Hyperparameters, projection details, and
constraint-validity checks are reported in
Appendix~\ref{app:hyperparameters},
Appendix~\ref{app:optimization_details}, and
Appendix~\ref{app:constraint_validity}.

%% file: Sections/results.tex
\section{Results}
\label{sec:results}

We evaluate whether explicit allocation guidance improves optimization within
the common capacity-limited threat model, then examine how the perturbation
budget is redistributed and which components of the mechanism account for
the observed gains. All main comparisons in this section refer to the
matched APGD/PMA-style baselines defined in the Experimental Setup;
specialized $\ell_1$ reference attacks are reported separately in
Appendix~\ref{app:cross_regime}.

\subsection{Main Performance under Capacity Limits}
\label{sec:main_results}

Table~\ref{tab:main_results} reports conditional attack success rate (ASR)
across the ten model--dataset configurations.

\begin{table*}[t]
\centering
\caption{\textbf{Conditional ASR (\%) under the common capacity-limited
threat model.} $\Delta_{\mathrm{best}}$ is Ours minus the highest-ASR
matched APGD/PMA baseline in the same configuration.}
\label{tab:main_results}
\small
\begin{tabular}{llrccccc}
\toprule
Dataset & Model & $N$
& APGD-CE
& APGD-DLR
& PMA
& Ours
& $\Delta_{\mathrm{best}}$ \\
\midrule
CIFAR-10
& WRN-94-16
& 9,368
& 8.64
& 10.28
& 9.84
& \textbf{17.98}
& \textbf{+7.70}
\\

CIFAR-10
& WRN-28-10
& 7,500
& 10.81
& 12.80
& 12.47
& \textbf{20.25}
& \textbf{+7.45}
\\

CIFAR-10
& XCiT-S12
& 9,006
& 15.08
& 17.03
& 17.03
& \textbf{22.52}
& \textbf{+5.49}
\\
\midrule

CIFAR-100
& WRN-28-10
& 7,383
& 22.54
& 28.44
& 27.59
& \textbf{35.41}
& \textbf{+6.97}
\\

CIFAR-100
& XCiT-S12
& 6,734
& 25.10
& 30.62
& 30.83
& \textbf{33.35}
& \textbf{+2.52}
\\

CIFAR-100
& WRN-70-16
& 7,522
& 20.15
& 25.38
& 24.82
& \textbf{32.53}
& \textbf{+7.15}
\\
\midrule

ImageNet
& ConvNeXt-L
& 1,000
& 24.90
& 24.70
& 27.00
& \textbf{42.90}
& \textbf{+15.90}
\\

ImageNet
& Peng et al.
& 1,000
& 35.10
& 34.80
& 37.20
& \textbf{54.90}
& \textbf{+17.70}
\\

ImageNet
& ViT-B-ConvStem
& 1,000
& 31.00
& 30.80
& 34.10
& \textbf{44.00}
& \textbf{+9.90}
\\

ImageNet
& Swin-L
& 1,000
& 25.80
& 27.30
& 28.20
& \textbf{36.00}
& \textbf{+7.80}
\\
\bottomrule
\end{tabular}
\end{table*}

Ours achieves higher ASR than all three matched baselines in every
configuration. Relative to the highest-ASR matched baseline in each row,
the improvement ranges from $2.52$ percentage points on
CIFAR-100/XCiT-S12 to $17.70$ points on ImageNet/Peng et al.

The pattern is not restricted to one dataset or architecture family.
Dataset-level macro-averages for Ours are $20.25\%$, $33.76\%$, and
$44.45\%$ on CIFAR-10, CIFAR-100, and ImageNet, respectively. The
corresponding strongest baseline macro-averages are $13.37\%$,
$28.15\%$, and $31.63\%$.

\begin{figure}[t]
    \centering
    \includegraphics[width=\linewidth]
    {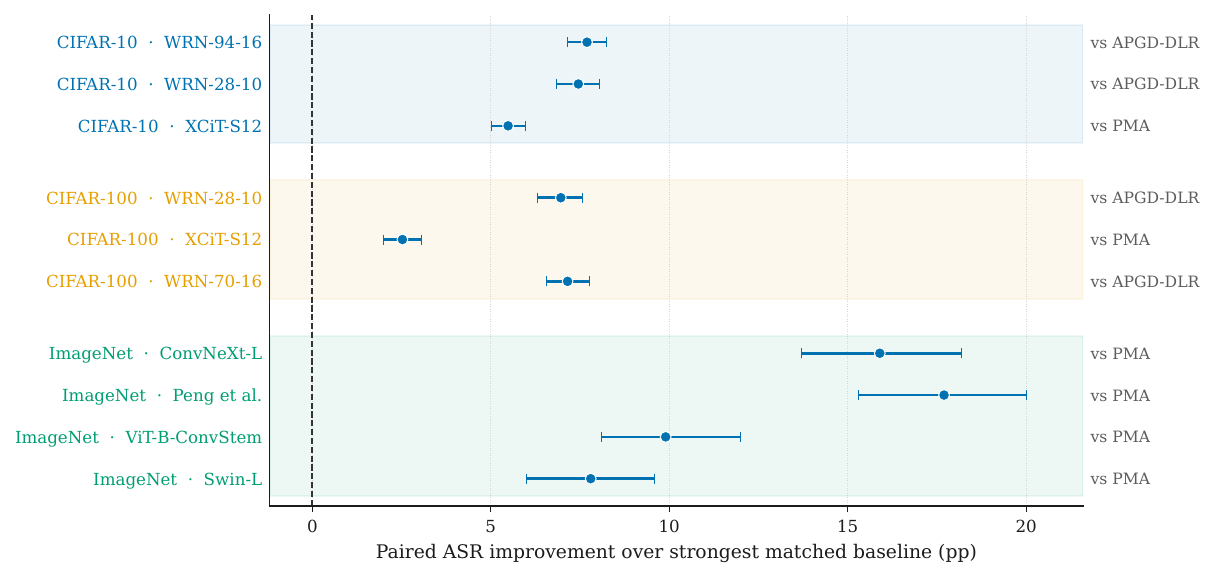}
    \caption{\textbf{Paired ASR improvement over the highest-ASR matched
    APGD/PMA baseline in each configuration.} Horizontal bars show 95\%
    paired confidence intervals; full statistics are reported in
    Appendix~\ref{app:statistics}.}
    \label{fig:paired_improvement}
\end{figure}

Figure~\ref{fig:paired_improvement} provides a compact view of the paired
effect in each configuration. All ten differences are positive and their
95\% paired confidence intervals remain above zero. Full pairwise
statistics and corrected McNemar tests are reported in
Appendix~\ref{app:statistics}.

\subsection{Where Does the Perturbation Budget Go?}
\label{sec:allocation_results}

The performance gain is accompanied by a clear change in where the shared
budget is spent. Table~\ref{tab:allocation} reports the fraction of total
absolute perturbation mass placed in the highest-importance $10\%$ of
spatial locations.

\begin{table*}[t]
\centering
\caption{\textbf{Perturbation mass (\%) in the highest-importance 10\% of
spatial locations.} Uniform spatial allocation corresponds to approximately
$10\%$.}
\label{tab:allocation}
\small
\begin{tabular}{llcccc}
\toprule
Dataset & Model
& APGD-CE
& APGD-DLR
& PMA
& Ours \\
\midrule
CIFAR-10  & WRN-94-16      & 11.12 & 10.84 & 11.05 & \textbf{30.36} \\
CIFAR-10  & WRN-28-10      & 11.09 & 10.83 & 11.05 & \textbf{29.14} \\
CIFAR-10  & XCiT-S12       & 10.72 & 10.61 & 10.69 & \textbf{25.19} \\
\midrule
CIFAR-100 & WRN-28-10      & 11.31 & 11.08 & 11.17 & \textbf{24.96} \\
CIFAR-100 & XCiT-S12       & 11.03 & 11.02 & 11.17 & \textbf{22.77} \\
CIFAR-100 & WRN-70-16      & 11.20 & 11.08 & 11.13 & \textbf{26.22} \\
\midrule
ImageNet  & ConvNeXt-L     & 10.81 & 10.72 & 10.66 & \textbf{21.60} \\
ImageNet  & Peng et al.    & 11.04 & 10.72 & 10.74 & \textbf{20.83} \\
ImageNet  & ViT-B-ConvStem & 10.61 & 10.56 & 10.56 & \textbf{19.41} \\
ImageNet  & Swin-L         & 11.08 & 10.76 & 10.91 & \textbf{19.15} \\
\bottomrule
\end{tabular}
\end{table*}

Matched baselines place approximately $10.6$--$11.3\%$ of their
perturbation mass in the top-$10\%$ importance region, close to uniform
spatial allocation. Ours increases this share to $19.15$--$30.36\%$.
The dataset-level averages are $28.23\%$ on CIFAR-10, $24.65\%$ on
CIFAR-100, and $20.25\%$ on ImageNet.

This redistribution does not arise from greater global budget consumption:
the realized $255\|\delta\|_1/D$ averages $3.91$, $3.96$, and $3.97$ on
CIFAR-10, CIFAR-100, and ImageNet, respectively, under the shared budget of
$4.0$.

\begin{figure}[t]
    \centering
    \includegraphics[width=\columnwidth]
    {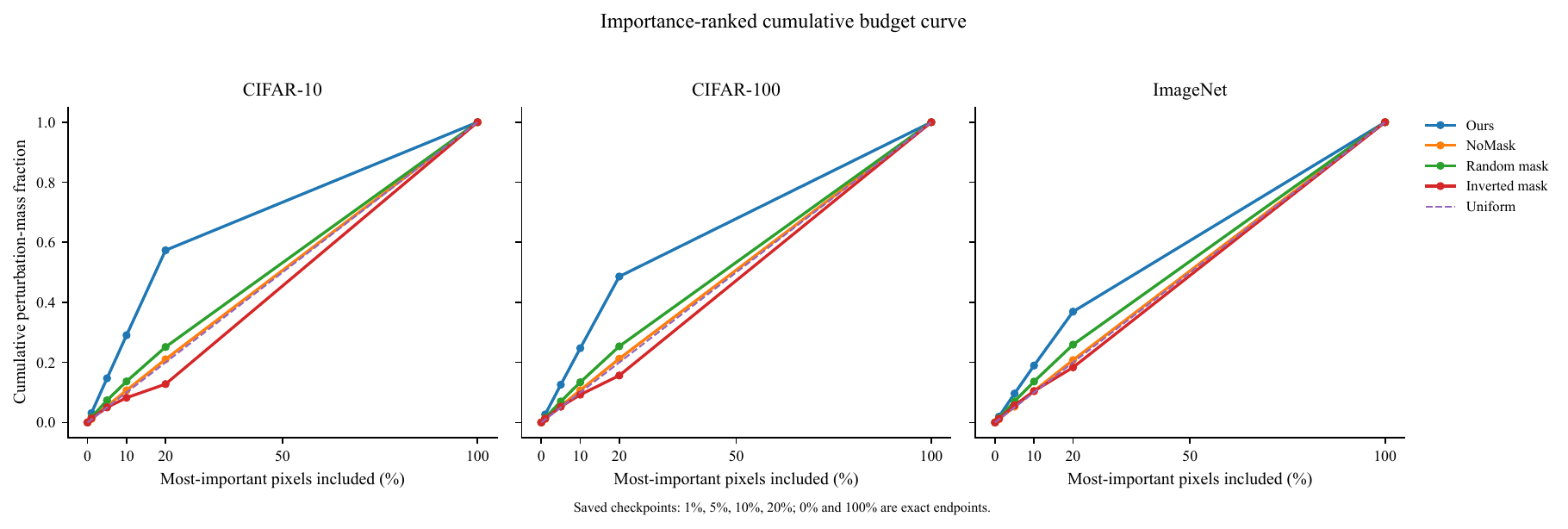}
    \caption{\textbf{Cumulative perturbation mass over importance-ranked
    locations.} Locations are ordered from highest to lowest clean-gradient
    importance; the dashed diagonal denotes uniform allocation.}
    \label{fig:allocation_curve}
\end{figure}

Figure~\ref{fig:allocation_curve} shows that the effect is not specific to
the Top-10 threshold. Perturbation mass accumulates more rapidly as
high-importance locations are included, indicating a systematic shift in
allocation rather than a threshold-specific statistic. Complementary
importance--perturbation correlation, effective-support, spatial-Gini, and
Top-1/5/20 diagnostics are reported in
Appendix~\ref{app:allocation_diagnostics}.

The same behavior is visible qualitatively in
Figure~\ref{fig:qualitative_allocation}. Under identical constraints,
changing or removing the allocation prior changes where perturbation
magnitude is concentrated relative to the fixed clean-gradient importance
map.

\subsection{Mechanism Ablations}
\label{sec:mechanism_ablations}

We next isolate the roles of centering, non-uniform redistribution, and the
quality of the importance prior. Table~\ref{tab:mechanism_ablation} compares
the full method with variants that remove, uncenter, randomize, invert, or
smooth the allocation signal.

\begin{table}[t]
\centering
\caption{\textbf{Mechanism ablations, conditional ASR (\%).}
\textsc{RandomMask} reports the mean over five spatial permutations.}
\label{tab:mechanism_ablation}
\small
\begin{tabular}{lccc}
\toprule
Variant & CIFAR-10 & CIFAR-100 & ImageNet \\
\midrule
NoMask       & 10.2 & 23.9 & 24.0 \\
Uncentered   &  3.9 & 13.7 & 20.4 \\
RandomMask   & 14.2 & 28.2 & 29.2 \\
InvertedMask & 14.5 & 25.9 & 23.4 \\
SmoothMask   & 17.6 & 34.4 & 36.2 \\
\textbf{Ours}& \textbf{19.6} & \textbf{37.9} & \textbf{36.8} \\
\bottomrule
\end{tabular}
\end{table}

Removing the allocation reward reduces ASR by $9.4$, $14.0$, and $12.8$
percentage points on CIFAR-10, CIFAR-100, and ImageNet, respectively.
The uncentered variant performs below even the no-allocation control on all
three datasets, supporting the reward--penalty role of centering rather
than a globally positive magnitude reward.

The randomized prior reveals a separate redistribution effect. Despite
containing no meaningful model-derived spatial ordering, it improves over
NoMask from $10.2\%$ to $14.2\%$ on CIFAR-10, $23.9\%$ to $28.2\%$ on
CIFAR-100, and $24.0\%$ to $29.2\%$ on ImageNet. The effect is stable
across the five independent spatial permutations reported in
Appendix~\ref{app:mechanism_ablations}.

Replacing this randomized preference with the model-derived prior provides
a further gain, reaching $19.6\%$, $37.9\%$, and $36.8\%$. The ablations
therefore support two complementary effects:
\emph{centered non-uniform redistribution} can assist optimization, while
\emph{importance alignment} provides additional benefit.

Inverting the prior weakens performance relative to Ours, while smoothing
preserves most of the gain. Thus, the mechanism does not appear to depend
only on isolated gradient spikes.

\begin{figure}[t]
    \centering
    \includegraphics[width=\columnwidth]
    {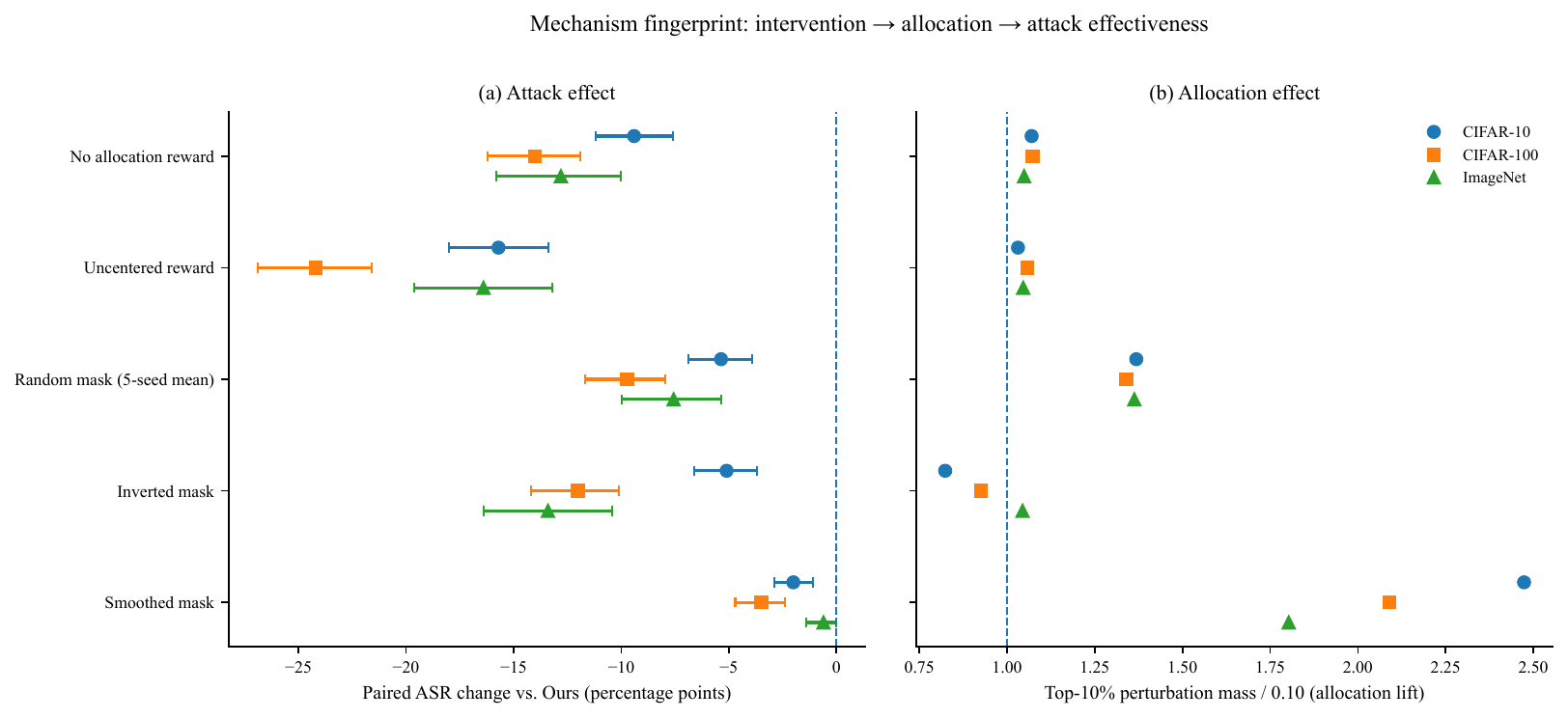}
    \caption{\textbf{Mechanism ablation: allocation versus attack effect.}
    Randomized guidance captures part of the benefit of non-uniform
    redistribution, while the model-derived prior provides an additional
    gain.}
    \label{fig:mechanism_fingerprint}
\end{figure}

Figure~\ref{fig:mechanism_fingerprint} connects these interventions to their
allocation behavior.

\subsection{Allocation Strength and Sensitivity}
\label{sec:sensitivity_results}

The allocation weight $\lambda$ trades off importance-aligned concentration
against optimization flexibility. On the CIFAR-10 reference sweep, ASR
rises from $10.2\%$ at $\lambda=0$ to a maximum of $20.6\%$ at
$\lambda=0.35$, then decreases to $14.7\%$ at $\lambda=1.0$, even as
importance concentration continues to increase.

The ImageNet diagnostic is also non-monotonic, and the value selected on
the independent calibration set is not the value that maximizes ASR on the
reported 500-image diagnostic population. We therefore retain the
preselected dataset-specific values rather than retuning on the evaluation
runs.

These results show that stronger concentration is not itself the objective:
the allocation term is most useful when it guides, rather than dominates,
the adversarial optimization. The complete CIFAR-10 and ImageNet
sensitivity curves and selection protocol are reported in
Appendix~\ref{app:lambda_selection}.

\subsection{Optimization Controls}
\label{sec:optimization_controls}

Several controls test alternative explanations for the observed gain.

First, the effect is not specific to cross-entropy. Adding the centered
allocation mechanism improves the corresponding unguided attack for all
four evaluated objectives---cross-entropy, margin, DLR, and probability
margin---on all three reference datasets. Moreover, the strongest guided
objective differs across datasets: margin reaches $23.1\%$ on CIFAR-10,
DLR reaches $42.0\%$ on CIFAR-100, and probability margin reaches
$43.8\%$ on ImageNet. This indicates that allocation guidance is
complementary to the choice of adversarial loss rather than specific to the
reference cross-entropy objective.

Second, compute matching does not explain the result. Replacing one of the
50 iterative attack steps with the clean-gradient computation yields the
same reference ASRs as the $50+1$ configuration:
$19.6\%$, $37.9\%$, and $36.8\%$ on CIFAR-10, CIFAR-100, and ImageNet.

Third, replacing the sequential feasibility operator with the exact
projection changes ASR only marginally; full results are reported in
Appendix~\ref{app:projection_control}.

Finally, on the CIFAR-10 and ImageNet reference controls, replacing the
heterogeneous capacity envelope with a uniform $12/255$ cap produces the
same ASR as the heterogeneous setting. This suggests that the observed
allocation gain in these controls is not specific to the sampled capacity
heterogeneity. Capacity-structure details are reported in
Appendix~\ref{app:capacity_details}, while the remaining mechanism controls
are reported in Appendix~\ref{app:mechanism_ablations}.

\subsection{Computational Cost and Perturbation Diagnostics}
\label{sec:cost_diagnostics}

Our method requires one additional backward evaluation to construct the
clean-gradient prior: $51$ backward evaluations versus $50$ for the matched
baselines. It uses $103$ forward evaluations, compared with $102$ for PMA
and $53$ for the APGD variants. Recorded wall-clock runtime is consequently
close to PMA, while APGD remains cheaper in forward-evaluation count.
Complete timing results are reported in Appendix~\ref{app:compute}.

The stronger allocation concentration also changes the local perturbation
profile. Although Ours does not consume a larger global mean-$\ell_1$
budget, it exhibits larger local magnitudes and greater perceptual
distortion than the more diffuse matched baselines. The same pattern is
observed in paired perceptual comparisons restricted to common-success
examples. We therefore interpret the performance gain as an allocation
effect, not as a simultaneous improvement in perceptual quality. Full
$\ell_1$, $\ell_\infty$, high-quantile magnitude, LPIPS, DISTS, and SSIM
diagnostics are reported in
Appendix~\ref{app:perceptual_diagnostics}.

%% file: Sections/limitations.tex
\section{Scope and Limitations}
\label{sec:limitations}

Our results support a specific claim about \emph{perturbation allocation}
within the evaluated capacity-limited threat model. The main comparisons use
matched APGD- and PMA-based baselines under the same global budget, local
capacities, image-range constraints, and evaluation samples. Within this
setting, importance-guided allocation improves attack success across all ten
model--dataset configurations. We do not claim universal superiority over
specialized $\ell_1$ attacks: adapted $\ell_1$-APGD remains stronger in the
reference comparisons in Appendix~\ref{app:cross_regime}.

The conventional and capacity-limited results in
Appendix~\ref{app:cross_regime} are descriptive rather than causal because
the settings differ in coordinate-wise capacities and global budget
parameterization. Our threat model represents one family of locally
constrained $\ell_1$ settings. Although a uniform-capacity control gives
similar results, other capacity structures may behave differently.

Our allocation prior is a fixed clean-gradient importance map computed once
before the attack, so the study does not establish clean gradients as the
optimal allocation signal. Dynamic, learned, or alternative estimators
remain possible extensions. The main experiments use 50 optimization steps
and one restart, and do not exhaust substantially larger compute budgets.

Finally, $\lambda$ is selected per dataset on a held-out calibration set and
then frozen across models. Stronger importance concentration does not
monotonically improve attack success, so the selected values are practical
operating points rather than globally optimal settings. The method can also
increase local perturbation magnitude and perceptual distortion while
remaining within the same global mean-$\ell_1$ budget; corresponding
diagnostics are reported in Appendix~\ref{app:perceptual_diagnostics}.

Overall, explicit allocation guidance can materially improve matched
adversarial optimization under the evaluated shared-budget, locally
constrained setting, without establishing universal superiority over
specialized $\ell_1$ attacks or the causal effect of local capacities across
separately parameterized threat regimes.

%% file: Sections/conclusion.tex
\section{Conclusion}
\label{sec:conclusion}

We studied adversarial optimization under a shared global mean-$\ell_1$
budget with coordinate-wise capacity constraints, focusing on the
\emph{allocation} of perturbation within a fixed feasible set.

We introduced a centered importance-guided allocation objective that uses a
fixed model-derived prior to steer perturbation toward more informative
regions without changing the admissible perturbation set. Across ten robust
model--dataset configurations, this mechanism improves attack success over
matched APGD- and PMA-based baselines under the same capacity-limited
constraints. Allocation diagnostics and ablations indicate that both
centered non-uniform redistribution and alignment with model-derived
importance contribute to the gain.

These results identify perturbation allocation as a distinct optimization
dimension under shared-budget, locally constrained threat models. They do
not imply universal superiority over specialized $\ell_1$ attacks, but
suggest that explicit allocation guidance can be a useful component of
future adversarial optimizers.

%% file: Appendix/full_threat_model_details.tex
\section{Full Threat-Model Details}
\label{app:threat_model}

This appendix specifies the threat model used throughout the main
capacity-limited evaluation and clarifies the roles of the global
$\ell_1$ budget, local coordinate-wise capacities, image-range constraints,
and model-derived importance. The central distinction is that the threat
model defines \emph{which perturbations are feasible}, whereas the
importance mechanism introduced in the Method section affects only
\emph{which feasible perturbations are preferred by the optimizer}.

\subsection{Capacity-Limited Feasible Set}
\label{app:threat_model_set}

Let
\[
x \in [0,1]^D
\]
denote a clean input that is correctly classified by the evaluated model,
and let
\[
\delta \in \mathbb{R}^{D}
\]
denote an additive adversarial perturbation. The capacity-limited feasible
set is
\begin{equation}
\mathcal{C}(x,E,b)
=
\left\{
\delta :
x+\delta \in [0,1]^D,\;
|\delta_i| \leq E_i,\;
\frac{\|\delta\|_1}{D} \leq b
\right\},
\label{eq:appendix_feasible_set}
\end{equation}
where:

\begin{itemize}
    \item $b$ is the global mean-$\ell_1$ perturbation budget;
    \item $E_i$ is the maximum admissible perturbation magnitude at
    coordinate $i$; and
    \item $x+\delta\in[0,1]^D$ enforces the valid image range.
\end{itemize}

The main experiments use
\begin{equation}
b=\frac{4}{255}.
\label{eq:appendix_main_b}
\end{equation}

The same feasible-set definition is used for all matched methods in the
main evaluation.

\subsection{Global Mean-$\ell_1$ Budget}
\label{app:mean_l1_details}

We parameterize the global constraint using the mean absolute perturbation
per scalar input coordinate,
\begin{equation}
\frac{\|\delta\|_1}{D}
\leq b.
\label{eq:mean_l1_constraint}
\end{equation}

Equivalently, the corresponding total $\ell_1$ radius is
\begin{equation}
\|\delta\|_1
\leq
Db.
\label{eq:total_l1_equivalent}
\end{equation}

Using a mean-$\ell_1$ parameterization makes the nominal budget independent
of the number of scalar input coordinates. Consequently, two datasets with
different image resolutions may have different equivalent total
$\ell_1$ radii even when they use the same value of $b$.

For an RGB image of height $H$ and width $W$,
\begin{equation}
D = 3HW,
\end{equation}
and therefore
\begin{equation}
\epsilon_1^{\mathrm{total}}
=
3HWb.
\label{eq:resolution_total_radius}
\end{equation}

Throughout the main capacity-limited experiments, comparisons are made
between attacks operating on the same images and under the same value of
$b$. Differences in attack success therefore cannot arise from one matched
method receiving a larger global mean-$\ell_1$ budget.

\subsection{Local Coordinate-Wise Capacities}
\label{app:local_capacity_details}

The local constraint
\begin{equation}
|\delta_i| \leq E_i
\label{eq:local_capacity_constraint}
\end{equation}
limits the amount of perturbation that may be assigned to an individual
coordinate. This restriction is related to previously studied sparse and
component-wise constrained threat models
\cite{croce2019sparseimperceivableadversarialattacks}, but in our formulation the capacity values do not
encode model importance.

The main experiments use a spatial capacity map with nominal parameters
\begin{equation}
\mu_E = \frac{12}{255},
\qquad
\sigma_E = \frac{2}{255}.
\label{eq:appendix_capacity_parameters}
\end{equation}

A single capacity value is generated for each spatial location and shared
across its RGB channels. Thus, if $p$ denotes a spatial location,
\begin{equation}
E_{1,p}
=
E_{2,p}
=
E_{3,p}
=
E_p.
\label{eq:capacity_rgb_sharing}
\end{equation}

The exact construction of $E_p$, including the truncated inverse-CDF
sampling procedure, is given in
Appendix~\ref{app:capacity_details}.

\subsection{Interaction Between Global and Local Constraints}
\label{app:global_local_interaction}

The global and local constraints play different roles. The global
mean-$\ell_1$ constraint limits the \emph{total} amount of perturbation that
may be distributed over the image, whereas $E_i$ limits how much of that
budget may be placed at any one coordinate.

Without local capacities, a large fraction of the shared $\ell_1$ budget
may in principle be concentrated on a comparatively small number of
coordinates. Under Equation~\ref{eq:appendix_feasible_set}, this is possible
only until those coordinates reach their corresponding capacities.

For example, if a coordinate reaches
\begin{equation}
|\delta_i| = E_i,
\end{equation}
additional perturbation magnitude cannot be assigned to that coordinate
while remaining feasible. Any remaining useful global budget must instead
be allocated among other coordinates.

Importantly, the local-capacity constraint does not necessarily reduce the
amount of global budget that can be consumed. When sufficiently many
coordinates remain below capacity, the full global budget $Db$ may still
be feasible. Rather, the local constraint changes the set of possible
\emph{distributions} of that budget across the input.

This distinction motivates our focus on perturbation allocation.

\subsection{Feasibility Is Independent of Importance}
\label{app:importance_feasibility_separation}

Let $M$ denote the clean-gradient importance map used by the proposed
method. The feasible set in
Equation~\ref{eq:appendix_feasible_set} contains no dependence on $M$:
\begin{equation}
\mathcal{C}
=
\mathcal{C}(x,E,b),
\qquad
\mathcal{C}
\neq
\mathcal{C}(x,E,b,M).
\label{eq:importance_independent_feasible_set}
\end{equation}

In particular, model-derived importance does not:

\begin{itemize}
    \item remove low-importance coordinates from the feasible set;
    \item increase the admissible magnitude at high-importance coordinates;
    \item decrease the local capacity of low-importance coordinates; or
    \item modify the global $\ell_1$ budget.
\end{itemize}

The importance map enters only through the allocation objective
\begin{equation}
J_{\mathrm{ours}}(\delta)
=
J_{\mathrm{adv}}(x+\delta,y)
+
\lambda R_{\mathrm{alloc}}(\delta,M),
\end{equation}
as described in the Centered Allocation Objective subsection.

This yields the separation
\begin{equation}
\underbrace{
x,\;E,\;b
}_{\text{define feasibility}}
\qquad\text{versus}\qquad
\underbrace{
M,\;\lambda
}_{\text{define allocation preference}}.
\label{eq:feasibility_preference_separation}
\end{equation}

This separation is essential to the interpretation of the main
experiments: matched baselines and our method receive the same admissible
perturbation region, while differing in how they optimize within it.

\subsection{Matched Threat-Model Protocol}
\label{app:matched_threat_protocol}

For every matched comparison in
the Main Performance under Capacity Limits subsection, attacks operate on:

\begin{itemize}
    \item the same clean input $x$ and ground-truth label $y$;
    \item the same global mean-$\ell_1$ budget $b$;
    \item the same realization of the local capacity map $E$;
    \item the same valid image domain $[0,1]^D$; and
    \item the same evaluation population.
\end{itemize}

Where the attack construction permits direct matching, the same initial
perturbation is also used. Baseline-specific optimization procedures are
preserved as described in
Appendix~\ref{app:baseline_adaptations}.

Constraint-satisfaction diagnostics confirming compliance with the global
budget, local capacities, and image range are reported in
Appendix~\ref{app:constraint_validity}.

\subsection{Relation to Conventional $\ell_1$ Constraints}
\label{app:standard_l1_relation}

For reference, a conventional $\ell_1$ threat model can be written as
\begin{equation}
\mathcal{C}_{\ell_1}(x,\epsilon_1)
=
\left\{
\delta:
x+\delta\in[0,1]^D,\;
\|\delta\|_1\leq\epsilon_1
\right\},
\label{eq:conventional_l1_set}
\end{equation}
without the additional coordinate-wise capacity constraint $E$.

The conventional-$\ell_1$ and capacity-limited evaluations included in
this work were originally configured using separately chosen global budget
parameterizations. They therefore differ in more than the presence or
absence of $E$. For this reason, we do not interpret their difference as a
controlled estimate of the isolated effect of local capacities.

Those results are retained only as a descriptive reference and are
reported separately in
Appendix~\ref{app:cross_regime}. All central empirical claims in the main
paper instead rely on matched comparisons within the common feasible set
of Equation~\ref{eq:appendix_feasible_set}.

%% file: Appendix/projection_and_optimization_details.tex
\section{Projection and Optimization Details}
\label{app:optimization_details}

This appendix provides additional details for the optimization procedure
introduced in the Attack Optimization subsection. The proposed method
optimizes the adversarial objective and allocation reward jointly while
restoring feasibility after every update.

\subsection{Initialization}

For each attacked input, the perturbation is initialized from a random
direction and projected into the capacity-limited feasible set
$\mathcal{C}(x,E,b)$. The main experiments use the same initial perturbation
for matched methods whenever the corresponding optimization procedure
permits direct initialization matching.

The recorded initialization mode for the proposed optimizer is
\texttt{random\_boundary}. The shared experiment seed is
\begin{equation}
s = 23.
\end{equation}

Using the same initialization and local-capacity realization across matched
methods reduces variation arising from stochastic starting points rather
than from the attack mechanism itself.

\subsection{Gradient Update}

At optimization iteration $t$, we compute
\begin{equation}
g_t
=
\nabla_{\delta}
J_{\mathrm{ours}}(\delta_t),
\end{equation}
where $J_{\mathrm{ours}}$ is defined in
Equation~\ref{eq:ours_objective}.

The gradient is normalized by its mean absolute magnitude,
\begin{equation}
\widetilde{g}_t
=
\frac{
g_t
}{
\operatorname{mean}(|g_t|)+\epsilon_g
},
\label{eq:appendix_gradient_normalization}
\end{equation}
and accumulated using momentum,
\begin{equation}
v_t
=
\mu v_{t-1}
+
\widetilde{g}_t,
\label{eq:appendix_momentum}
\end{equation}
with
\begin{equation}
\mu = 0.75.
\end{equation}

The perturbation is updated in the sign direction of the accumulated
gradient,
\begin{equation}
\widetilde{\delta}_{t+1}
=
\delta_t
+
\alpha_t
\operatorname{sign}(v_t),
\label{eq:appendix_update}
\end{equation}
where $\alpha_t$ follows the cosine-decay schedule used in the attack
implementation. The recorded reference step scale is
\begin{equation}
\alpha_{\mathrm{ref}}
=
\frac{2}{255}.
\end{equation}

All main experiments use 50 optimization steps and one restart unless
otherwise stated.

\subsection{Sequential Feasibility Operator}

After each unconstrained update, feasibility is restored in three stages.
First, the perturbation is clipped to the coordinate-wise capacities:
\begin{equation}
\delta_i
\leftarrow
\operatorname{clip}
\left(
\widetilde{\delta}_i,
-E_i,
E_i
\right).
\label{eq:appendix_cap_clip}
\end{equation}

Second, the valid image range is enforced:
\begin{equation}
\delta
\leftarrow
\operatorname{clip}(x+\delta,0,1)-x.
\label{eq:appendix_image_clip}
\end{equation}

Finally, if necessary, the perturbation is projected onto the global
$\ell_1$ ball of radius $Db$:
\begin{equation}
\delta
\leftarrow
\Pi_{\|\cdot\|_1\leq Db}(\delta),
\label{eq:appendix_l1_projection}
\end{equation}
using the Euclidean projection algorithm of
\cite{duchi2008efficient}.

We denote the complete sequential operator by
\begin{equation}
\Pi_{\mathrm{seq}}(\widetilde{\delta};x,E,b).
\end{equation}

Because the final $\ell_1$ projection only reduces perturbation magnitude,
it does not introduce new violations of the previously enforced local or
image-range constraints.

\subsection{Exact-Projection Control}
\label{app:projection_control}

To test whether the main results depend on the sequential feasibility
operator, we additionally compare it with direct Euclidean projection onto
the full intersection
\begin{equation}
\Pi_{\mathcal{C}}(z)
=
\arg\min_{\delta\in\mathcal{C}(x,E,b)}
\frac{1}{2}
\|\delta-z\|_2^2.
\label{eq:exact_projection}
\end{equation}

Table~\ref{tab:projection_control} shows that replacing the sequential
operator with the exact projection changes ASR only slightly on the three
reference datasets.

\begin{table}[t]
\centering
\caption{
Projection-operator control. Values are conditional ASR (\%) on the
reference ablation populations.
}
\label{tab:projection_control}
\small
\begin{tabular}{lccc}
\toprule
Projection & CIFAR-10 & CIFAR-100 & ImageNet \\
\midrule
Sequential & 19.6 & 37.9 & 36.8 \\
Exact      & 19.4 & 37.7 & 37.2 \\
\bottomrule
\end{tabular}
\end{table}

The corresponding differences are $-0.2$, $-0.2$, and $+0.4$ percentage
points, respectively. This control indicates that the observed allocation
effect is not driven by the particular sequential projection used in the
main experiments.

\subsection{Best-Candidate Tracking}

Optimization does not rely only on the final iterate. At every iteration
and restart, feasible candidates are ranked by the probability margin
\begin{equation}
m_{\mathrm{prob}}(x+\delta)
=
\max_{k\neq y} p_k(x+\delta)-p_y(x+\delta),
\end{equation}
and the candidate with the largest observed margin is retained.

This ranking criterion is used only for candidate selection. Unless
explicitly varied in an objective ablation, the optimization gradient
continues to follow the adversarial objective together with the allocation
reward.

%% file: Appendix/capacity_map_construction.tex
\section{Capacity-Map Construction}
\label{app:capacity_details}

The main threat model assigns a local perturbation capacity to each spatial
location. These capacities are generated independently of the importance
map used by the proposed allocation mechanism.

\subsection{Sampling Procedure}

For spatial location $p$, we sample
\begin{equation}
U_p
\sim
\mathcal{U}(u_{\min},u_{\max}),
\end{equation}
with
\begin{equation}
u_{\min}=10^{-3},
\qquad
u_{\max}=1-10^{-3}.
\end{equation}

The spatial capacity is then
\begin{equation}
E_p
=
\mu_E
+
\sigma_E
\Phi^{-1}(U_p),
\label{eq:appendix_capacity_map}
\end{equation}
where $\Phi^{-1}$ is the inverse standard-normal CDF.

The main setting uses
\begin{equation}
\mu_E=\frac{12}{255},
\qquad
\sigma_E=\frac{2}{255}.
\end{equation}

Each sampled spatial capacity is shared across the three RGB channels:
\begin{equation}
E_{1,p}=E_{2,p}=E_{3,p}=E_p.
\end{equation}

The truncation of $U_p$ avoids extreme Gaussian tails while retaining
moderate spatial variation around the nominal $12/255$ local scale.

\subsection{Independence from Importance}

The capacity-generation process does not use gradients, saliency values,
labels, or the allocation map. Consequently,
\begin{equation}
E
\perp
M
\quad
\text{by construction},
\end{equation}
in the sense that the capacity realization is generated independently of
the model-derived importance prior.

The same realization is supplied to all matched attacks for a given
sample. Digest-based consistency checks for the shared capacity maps are
reported in Appendix~\ref{app:constraint_validity}.

\subsection{Capacity-Structure Control}

We also evaluate whether the main result depends specifically on
heterogeneous capacities. The reference heterogeneous map is compared with
a uniform cap
\begin{equation}
E_p=\frac{12}{255}
\qquad
\forall p,
\end{equation}
and with a no-local-cap control.

\begin{table}[t]
\centering
\caption{
Local-capacity structure control. Values are conditional ASR (\%).
The heterogeneous envelope is the main setting.
}
\label{tab:capacity_structure}
\small
\begin{tabular}{lcc}
\toprule
Capacity structure & CIFAR-10 & ImageNet \\
\midrule
Heterogeneous envelope & 19.6 & 36.8 \\
Uniform $12/255$ cap   & 19.6 & 36.8 \\
No local cap           & 18.3 & 40.4 \\
\bottomrule
\end{tabular}
\end{table}

The identical ASR obtained by the heterogeneous and uniform-cap variants in
these reference runs indicates that the main allocation gain does not
depend on the Gaussian heterogeneity itself. We therefore treat the
heterogeneous map as one evaluated capacity model rather than as a necessary
component of the allocation mechanism.

%% file: Appendix/baseline_adaptation_details.tex
\section{Baseline Adaptation Details}
\label{app:baseline_adaptations}

All main-paper baselines operate under the same capacity-limited feasible
set as the proposed method. The adaptation is intended to preserve the
characteristic optimization rule or adversarial objective of each baseline
while enforcing the shared global budget, local capacities, and image
bounds.

\subsection{APGD-CE and APGD-DLR}

APGD-CE and APGD-DLR retain the corresponding APGD optimization objectives
and adaptive optimization structure
\cite{croce2020reliable}. Their candidate perturbations are constrained to
\begin{equation}
\mathcal{C}(x,E,b)
\end{equation}
using the common feasibility operator.

APGD-CE optimizes cross-entropy, whereas APGD-DLR uses the
difference-of-logits-ratio objective. Both main-paper variants use 50
optimization steps and one restart.

\subsection{Probability Margin Attack}

PMA retains its probability-margin adversarial objective
\cite{xie2025million} while operating under the same
$\mathcal{C}(x,E,b)$ constraint.

The main evaluation uses 50 optimization steps and one restart. As with
APGD, the adaptation concerns feasibility rather than introducing the
importance prior used by our method.

\subsection{Shared Assets}

For the matched APGD/PMA comparisons, the following quantities are shared
within each model--sample pair:

\begin{itemize}
    \item the clean image and label;
    \item the global mean-$\ell_1$ budget;
    \item the local-capacity realization;
    \item the random seed;
    \item the evaluation population; and
    \item the initial perturbation where direct matching is applicable.
\end{itemize}

The main result therefore compares different optimization strategies within
a common feasible region.

\subsection{Specialized $\ell_1$ Reference Attacks}

We additionally evaluate $\ell_1$-APGD
\cite{croce2021robustbenchstandardizedadversarialrobustness} and FMN-$\ell_1$
\cite{pintor2021fastminimumnormadversarialattacks} in the descriptive cross-regime experiment of
Appendix~\ref{app:cross_regime}.

The conventional $\ell_1$-APGD runs use 50 iterations, one restart, and
the standard larger-epsilon initialization option used in the underlying
implementation. Their total $\ell_1$ radii are $12$ on CIFAR and $60$ on
ImageNet.

The conventional FMN runs use 50 steps. The recorded configuration uses
maximum step size $1.0$, $\gamma=0.05$, and Foolbox version 3.3.4. Their
nominal total $\ell_1$ bounds are again $12$ on CIFAR and $60$ on
ImageNet.

For the capacity-limited reference, adapted versions enforce
\begin{equation}
\frac{\|\delta\|_1}{D}\leq\frac{4}{255}
\end{equation}
together with the same heterogeneous local-capacity envelope used by the
main experiments.

These specialized attacks are retained to document their behavior under the
separately configured settings. Because the conventional and
capacity-limited experiments do not use identical global-budget
parameterizations, they are not treated as controlled local-cap
interventions.

%% file: Appendix/dataset_model_evaluation_populations.tex
\section{Dataset, Model, and Evaluation Populations}
\label{app:evaluation_populations}

Table~\ref{tab:full_evaluation_populations} reports the complete evaluation
populations used in the main matched comparison.

\begin{table*}[t]
\centering
\caption{
Main evaluation populations. Clean accuracy and clean-correct counts are
computed on the full corresponding test or validation set. ``Evaluated''
denotes the number of clean-correct images included in the reported attack
run.
}
\label{tab:full_evaluation_populations}
\small
\begin{tabular}{llrrrrl}
\toprule
Dataset & Model
& Dataset size
& Clean acc. (\%)
& Clean-correct
& Evaluated
& Coverage \\
\midrule
CIFAR-10
& WRN-94-16
& 10,000 & 93.68 & 9,368 & 9,368 & complete \\
CIFAR-10
& WRN-28-10
& 10,000 & 92.16 & 9,216 & 7,500 & partial \\
CIFAR-10
& XCiT-S12
& 10,000 & 90.06 & 9,006 & 9,006 & complete \\
\midrule
CIFAR-100
& WRN-28-10
& 10,000 & 73.83 & 7,383 & 7,383 & complete \\
CIFAR-100
& XCiT-S12
& 10,000 & 67.34 & 6,734 & 6,734 & complete \\
CIFAR-100
& WRN-70-16
& 10,000 & 75.22 & 7,522 & 7,522 & complete \\
\midrule
ImageNet
& ConvNeXt-L
& 50,000 & 78.114 & 39,057 & 1,000 & fixed subset \\
ImageNet
& Peng et al.
& 50,000 & 73.882 & 36,941 & 1,000 & fixed subset \\
ImageNet
& ViT-B-ConvStem
& 50,000 & 76.718 & 38,359 & 1,000 & fixed subset \\
ImageNet
& Swin-L
& 50,000 & 79.180 & 39,590 & 1,000 & fixed subset \\
\bottomrule
\end{tabular}
\end{table*}

For CIFAR-10 and CIFAR-100, the intended population consists of
clean-correct examples from the complete 10,000-image test set. The
CIFAR-10 WRN-28-10 main run contains 7,500 of the 9,216 clean-correct
examples and is therefore explicitly reported as partial coverage rather
than as a complete-test evaluation.

For ImageNet, each model is evaluated on 1,000 clean-correct validation
images selected using the fixed seeded ordering recorded by the evaluation
pipeline. Methods compared within a given model use the same 1,000 images.

\subsection{Reference Populations for Ablations}

The mechanism ablations use fixed reference populations rather than all ten
main configurations.

\begin{table}[t]
\centering
\caption{
Reference populations used for the ablation studies.
}
\label{tab:ablation_populations}
\small
\begin{tabular}{llr}
\toprule
Dataset & Model & $N$ \\
\midrule
CIFAR-10  & WRN-28-10 & 1,000 \\
CIFAR-100 & WRN-28-10 & 1,000 \\
ImageNet  & Swin-L    & 500 \\
\bottomrule
\end{tabular}
\end{table}

Within an ablation suite, all variants operate on the same samples and
shared attack assets.

%% file: Appendix/complete_hyperparameters.tex
\section{Complete Hyperparameters}
\label{app:hyperparameters}

Table~\ref{tab:ours_hyperparameters} summarizes the reference configuration
of the proposed optimizer as recorded by the experiment artifacts.

\begin{table}[t]
\centering
\caption{
Reference hyperparameters for the proposed allocation-guided optimizer.
}
\label{tab:ours_hyperparameters}
\small
\begin{tabular}{ll}
\toprule
Parameter & Value \\
\midrule
Global mean-$\ell_1$ budget $b$ & $4/255$ \\
Capacity mean $\mu_E$ & $12/255$ \\
Capacity std. $\sigma_E$ & $2/255$ \\
Capacity mode & heterogeneous envelope \\
Capacity quantile range & $[10^{-3},1-10^{-3}]$ \\
Optimization steps & 50 \\
Restarts & 1 \\
Initialization & random boundary / shared initial $\delta$ \\
Reference step scale & $2/255$ \\
Momentum $\mu$ & $0.75$ \\
Reference objective & cross-entropy \\
Projection & sequential \\
Saliency source & true-class logit \\
Importance clipping quantile $q$ & $0.995$ \\
Importance exponent $\gamma$ & $1.0$ \\
Importance floor $M_{\min}$ & $0.05$ \\
Importance ceiling $M_{\max}$ & $8.0$ \\
Smooth absolute-value $\tau$ & $0.25/255$ \\
Reference smoothing $\sigma$ & $0$ \\
Candidate selection & probability margin \\
Experiment seed & 23 \\
\bottomrule
\end{tabular}
\end{table}

The dataset-specific allocation strengths are

\begin{equation}
\lambda =
\begin{cases}
0.5, & \text{CIFAR-10},\\
0.5, & \text{CIFAR-100},\\
0.2, & \text{ImageNet}.
\end{cases}
\end{equation}

The main APGD-CE, APGD-DLR, and PMA comparisons use 50 optimization steps
and one restart under the same threat-model parameters.

\subsection{Software Environment}

The ablation result records correspond to the following software
environment:

\begin{table}[t]
\centering
\caption{Recorded experimental software environment.}
\label{tab:software_environment}
\small
\begin{tabular}{ll}
\toprule
Component & Version / configuration \\
\midrule
Python & 3.12.13 \\
PyTorch & 2.10.0+cu128 \\
Torchvision & 0.25.0+cu128 \\
CUDA & 12.8 \\
cuDNN & 91002 \\
RobustBench & 1.1 \\
LPIPS & 0.1.4 \\
PIQ & 0.8.0 \\
GPU & 2 $\times$ Tesla T4 \\
\bottomrule
\end{tabular}
\end{table}


%% file: Appendix/lambda_selection_and_sensitivity.tex
\section{Lambda Selection and Sensitivity}
\label{app:lambda_selection}

The allocation weight $\lambda$ determines the relative contribution of
$R_{\mathrm{alloc}}$ to the attack objective. We select one value per
dataset on 100 held-out clean-correct examples and freeze it across models
within that dataset.

The candidate set used for selection is
\begin{equation}
\lambda
\in
\{0.10,0.20,0.35,0.50,0.75\}.
\end{equation}

The selected values are
\begin{equation}
\lambda_{\mathrm{C10}}=0.5,
\qquad
\lambda_{\mathrm{C100}}=0.5,
\qquad
\lambda_{\mathrm{ImageNet}}=0.2.
\end{equation}

The selected values are not retuned using the reported main-test results.

\subsection{Sensitivity on the Reference Populations}

The retained sensitivity runs additionally include $\lambda=0$ and
$\lambda=1$ as diagnostic endpoints.

\begin{table*}[t]
\centering
\caption{
Lambda sensitivity on the available reference runs. Top-10 mass is the
fraction of perturbation magnitude assigned to the highest-importance
10\% of spatial locations.
}
\label{tab:lambda_sensitivity}
\small
\begin{tabular}{lrrrrrrrr}
\toprule
& \multicolumn{2}{c}{$\lambda=0$}
& \multicolumn{2}{c}{$0.1$}
& \multicolumn{2}{c}{$0.2$}
& \multicolumn{2}{c}{$0.35$} \\
Dataset
& ASR & Top-10
& ASR & Top-10
& ASR & Top-10
& ASR & Top-10 \\
\midrule
CIFAR-10
& 10.2 & 10.70
& 17.9 & 24.27
& 20.3 & 26.57
& 20.6 & 28.21 \\
ImageNet
& 24.0 & 10.49
& 32.6 & 16.35
& 36.8 & 18.95
& 40.8 & 21.18 \\
\bottomrule
\end{tabular}

\vspace{0.5em}

\begin{tabular}{lrrrrrr}
\toprule
& \multicolumn{2}{c}{$\lambda=0.5$}
& \multicolumn{2}{c}{$0.75$}
& \multicolumn{2}{c}{$1.0$} \\
Dataset
& ASR & Top-10
& ASR & Top-10
& ASR & Top-10 \\
\midrule
CIFAR-10
& 19.6 & 29.06
& 16.9 & 29.81
& 14.7 & 30.24 \\
ImageNet
& 41.2 & 22.66
& 43.0 & 24.49
& 41.0 & 25.71 \\
\bottomrule
\end{tabular}
\end{table*}

The CIFAR-10 curve illustrates the central trade-off particularly clearly.
Increasing $\lambda$ initially improves both importance-aligned allocation
and ASR, but after an intermediate optimum the allocation concentration
continues to increase while attack success decreases.

The ImageNet diagnostic shows the same non-monotonic relationship between
allocation concentration and effectiveness, although the value selected on
the independent calibration set is not the value that maximizes ASR on this
500-image diagnostic population. We therefore retain the preselected
$\lambda=0.2$ rather than retrospectively tuning on the reported
evaluation subset.

This behavior supports the interpretation that the allocation term should
guide, rather than dominate, adversarial optimization.

\begin{figure*}[t]
    \centering
    \includegraphics[width=\textwidth]
    {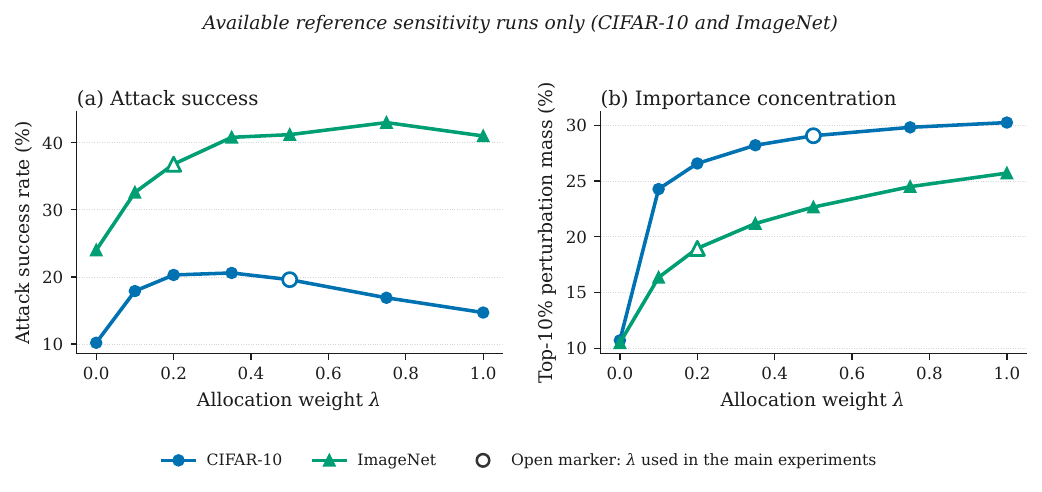}
    \caption{
    Lambda sensitivity. Left: ASR as a function of allocation weight.
    Right: perturbation mass assigned to the top-$10\%$ importance region.
    Increasing $\lambda$ produces progressively stronger importance
    concentration, but attack effectiveness is non-monotonic.
    }
    \label{fig:appendix_lambda_tradeoff}
\end{figure*}


%% file: Appendix/full_mechanism_ablations.tex
\section{Full Mechanism Ablations}
\label{app:mechanism_ablations}

This section reports the complete mechanism controls underlying
the Mechanism Ablations subsection. All variants within an ablation
suite use matched samples, threat constraints, and shared assets.

\subsection{Importance Information and Centering}

The core variants are:

\begin{itemize}
    \item \textbf{NoMask}: removes the allocation reward
    ($\lambda=0$);
    \item \textbf{Uncentered}: replaces the centered preference
    $M-1$ with positive importance weighting;
    \item \textbf{RandomMask}: replaces model-derived importance with a
    randomized spatial prior; five random seeds are evaluated;
    \item \textbf{InvertedMask}: reverses the importance ordering;
    \item \textbf{SmoothMask}: spatially smooths the importance prior; and
    \item \textbf{Ours}: uses the centered raw model-derived importance map.
\end{itemize}

\begin{table*}[t]
\centering
\caption{
Core mechanism ablations. ASR is in percent; Top-10 is perturbation-mass
share in the highest-importance 10\% of locations. RandomMask reports the
mean over five randomized priors, with the ASR standard deviation in
parentheses.
}
\label{tab:full_core_ablation}
\small
\begin{tabular}{lrrrrrr}
\toprule
& \multicolumn{2}{c}{CIFAR-10}
& \multicolumn{2}{c}{CIFAR-100}
& \multicolumn{2}{c}{ImageNet} \\
Variant
& ASR & Top-10
& ASR & Top-10
& ASR & Top-10 \\
\midrule
NoMask
& 10.2 & 10.70
& 23.9 & 10.74
& 24.0 & 10.49 \\
Uncentered
& 3.9 & 10.31
& 13.7 & 10.58
& 20.4 & 10.46 \\
RandomMask
& 14.24 (0.19) & 13.68
& 28.18 (0.16) & 13.40
& 29.24 (0.09) & 13.63 \\
InvertedMask
& 14.5 & 8.24
& 25.9 & 9.26
& 23.4 & 10.44 \\
SmoothMask
& 17.6 & 24.74
& 34.4 & 20.90
& 36.2 & 18.03 \\
\textbf{Ours}
& \textbf{19.6} & \textbf{29.06}
& \textbf{37.9} & \textbf{24.74}
& \textbf{36.8} & \textbf{18.95} \\
\bottomrule
\end{tabular}
\end{table*}

The randomized prior consistently improves over NoMask, demonstrating that
centered non-uniform redistribution itself contributes to optimization.
The true importance prior provides an additional improvement on all three
datasets.

The Uncentered variant performs below NoMask in all three cases, supporting
the use of a reward--penalty formulation rather than a globally positive
magnitude reward.

\subsection{Alternative Adversarial Objectives}

We next test whether the allocation gain is tied specifically to
cross-entropy.

\begin{table*}[t]
\centering
\caption{
Objective ablation. Each pair compares an objective without allocation
against the same objective augmented by the centered raw importance prior.
Values are conditional ASR (\%).
}
\label{tab:objective_ablation}
\small
\begin{tabular}{lrrrrrrrr}
\toprule
& \multicolumn{2}{c}{CE}
& \multicolumn{2}{c}{Margin}
& \multicolumn{2}{c}{DLR}
& \multicolumn{2}{c}{Prob. Margin} \\
Dataset
& NoMask & Guided
& NoMask & Guided
& NoMask & Guided
& NoMask & Guided \\
\midrule
CIFAR-10
& 10.2 & 19.6
& 11.7 & 23.1
& 11.9 & 22.3
& 11.8 & 18.6 \\
CIFAR-100
& 23.9 & 37.9
& 28.0 & 41.5
& 29.5 & 42.0
& 28.5 & 38.2 \\
ImageNet
& 24.0 & 36.8
& 24.0 & 30.2
& 24.4 & 39.2
& 26.6 & 43.8 \\
\bottomrule
\end{tabular}
\end{table*}

Allocation guidance improves the corresponding unguided variant for every
objective and dataset in this control.

\subsection{Compute-Matched Control}

The reference method uses 50 iterative attack steps plus one clean-gradient
computation. To test whether the extra backward evaluation explains the
gain, we replace one iterative step with the clean-gradient computation.

\begin{table}[t]
\centering
\caption{
Compute-matched control. ``Raw 49+1'' uses 49 iterative attack gradients
plus one clean-gradient computation.
}
\label{tab:compute_match_ablation}
\small
\begin{tabular}{lccc}
\toprule
Variant & CIFAR-10 & CIFAR-100 & ImageNet \\
\midrule
NoMask, 50 attack steps & 10.2 & 23.9 & 24.0 \\
Ours, 50 + 1            & 19.6 & 37.9 & 36.8 \\
Ours, 49 + 1            & 19.6 & 37.9 & 36.8 \\
\bottomrule
\end{tabular}
\end{table}

The identical ASRs of the $50+1$ and $49+1$ guided variants indicate that
the improvement cannot be explained by receiving one additional backward
evaluation.

\subsection{Optimization Controls}
\label{app:optimization_controls}

Two further controls test projection choice and local-capacity structure.

\begin{table}[t]
\centering
\caption{Projection control, conditional ASR (\%).}
\label{tab:appendix_projection_ablation}
\small
\begin{tabular}{lccc}
\toprule
Variant & CIFAR-10 & CIFAR-100 & ImageNet \\
\midrule
Sequential projection & 19.6 & 37.9 & 36.8 \\
Exact projection      & 19.4 & 37.7 & 37.2 \\
\bottomrule
\end{tabular}
\end{table}

\begin{table}[t]
\centering
\caption{Local-capacity structure control, conditional ASR (\%).}
\label{tab:appendix_cap_ablation}
\small
\begin{tabular}{lcc}
\toprule
Variant & CIFAR-10 & ImageNet \\
\midrule
Heterogeneous envelope & 19.6 & 36.8 \\
Uniform $12/255$ cap   & 19.6 & 36.8 \\
No local cap           & 18.3 & 40.4 \\
\bottomrule
\end{tabular}
\end{table}

The uniform-cap result closely matches the heterogeneous reference. The
no-cap control should be interpreted as an optimization diagnostic rather
than as the controlled cross-regime experiment discussed in
Appendix~\ref{app:cross_regime}.

%% file: Appendix/allocation_diagnostics.tex
\section{Allocation Diagnostics}
\label{app:allocation_diagnostics}

We analyze allocation using importance-ranked perturbation mass,
importance--perturbation correlation, effective spatial support, and spatial
Gini concentration.

\subsection{Main-Experiment Allocation Statistics}

\begin{table*}[t]
\centering
\caption{
Dataset-level macro-averaged allocation diagnostics. Top-10 is the fraction
of absolute perturbation mass in the highest-importance 10\% of spatial
locations.
}
\label{tab:allocation_diagnostics_macro}
\small
\begin{tabular}{llrrrr}
\toprule
Dataset & Method
& Top-10 (\%)
& Importance corr.
& Effective spatial frac.
& Spatial Gini \\
\midrule
CIFAR-10 & APGD-CE  & 10.98 & 0.096 & 0.842 & 0.244 \\
         & APGD-DLR & 10.76 & 0.064 & 0.815 & 0.270 \\
         & PMA      & 10.93 & 0.088 & 0.842 & 0.245 \\
         & Ours     & \textbf{28.23} & \textbf{0.692} & 0.403 & 0.623 \\
\midrule
CIFAR-100 & APGD-CE  & 11.18 & 0.121 & 0.839 & 0.247 \\
          & APGD-DLR & 11.06 & 0.086 & 0.792 & 0.290 \\
          & PMA      & 11.16 & 0.111 & 0.831 & 0.255 \\
          & Ours     & \textbf{24.65} & \textbf{0.665} & 0.471 & 0.575 \\
\midrule
ImageNet & APGD-CE  & 10.89 & 0.084 & 0.822 & 0.264 \\
         & APGD-DLR & 10.69 & 0.056 & 0.783 & 0.298 \\
         & PMA      & 10.72 & 0.069 & 0.827 & 0.259 \\
         & Ours     & \textbf{20.25} & \textbf{0.499} & 0.560 & 0.495 \\
\bottomrule
\end{tabular}
\end{table*}

The increased Top-10 share and importance correlation directly reflect the
intended allocation behavior. The reduced effective spatial fraction and
larger spatial Gini additionally show that the resulting perturbations are
less spatially diffuse.

\subsection{Cumulative Importance-Ranked Mass}

Table~\ref{tab:cumulative_allocation} reports cumulative perturbation mass
at the $1\%$, $5\%$, $10\%$, and $20\%$ importance checkpoints for the
reference mechanism variants.

\begin{table*}[t]
\centering
\caption{
Cumulative perturbation-mass share (\%) after including the most-important
spatial locations. A uniform allocation would yield approximately
$1\%$, $5\%$, $10\%$, and $20\%$ at the corresponding checkpoints.
}
\label{tab:cumulative_allocation}
\small
\begin{tabular}{llrrrr}
\toprule
Dataset & Variant & Top-1 & Top-5 & Top-10 & Top-20 \\
\midrule
CIFAR-10
& NoMask       & 1.16 & 5.44 & 10.70 & 21.05 \\
& InvertedMask & 1.35 & 5.03 &  8.24 & 12.78 \\
& SmoothMask   & 2.71 & 12.69 & 24.74 & 46.23 \\
& Ours         & \textbf{3.12} & \textbf{14.72}
               & \textbf{29.06} & \textbf{57.35} \\
\midrule
CIFAR-100
& NoMask       & 1.16 & 5.44 & 10.74 & 21.24 \\
& InvertedMask & 1.29 & 5.24 &  9.26 & 15.64 \\
& SmoothMask   & 2.27 & 10.67 & 20.90 & 40.11 \\
& Ours         & \textbf{2.67} & \textbf{12.58}
               & \textbf{24.74} & \textbf{48.61} \\
\midrule
ImageNet
& NoMask       & 1.07 & 5.28 & 10.49 & 20.79 \\
& InvertedMask & 1.26 & 5.70 & 10.44 & 18.35 \\
& SmoothMask   & 1.84 & 9.13 & 18.03 & 35.13 \\
& Ours         & \textbf{1.93} & \textbf{9.62}
               & \textbf{18.95} & \textbf{36.91} \\
\bottomrule
\end{tabular}
\end{table*}

The cumulative curves therefore show a graded rather than purely binary
change in allocation: the difference is already visible in the most
important few percent of locations and persists through larger importance
regions.

\begin{figure*}[t]
    \centering
    \includegraphics[width=\textwidth]
    {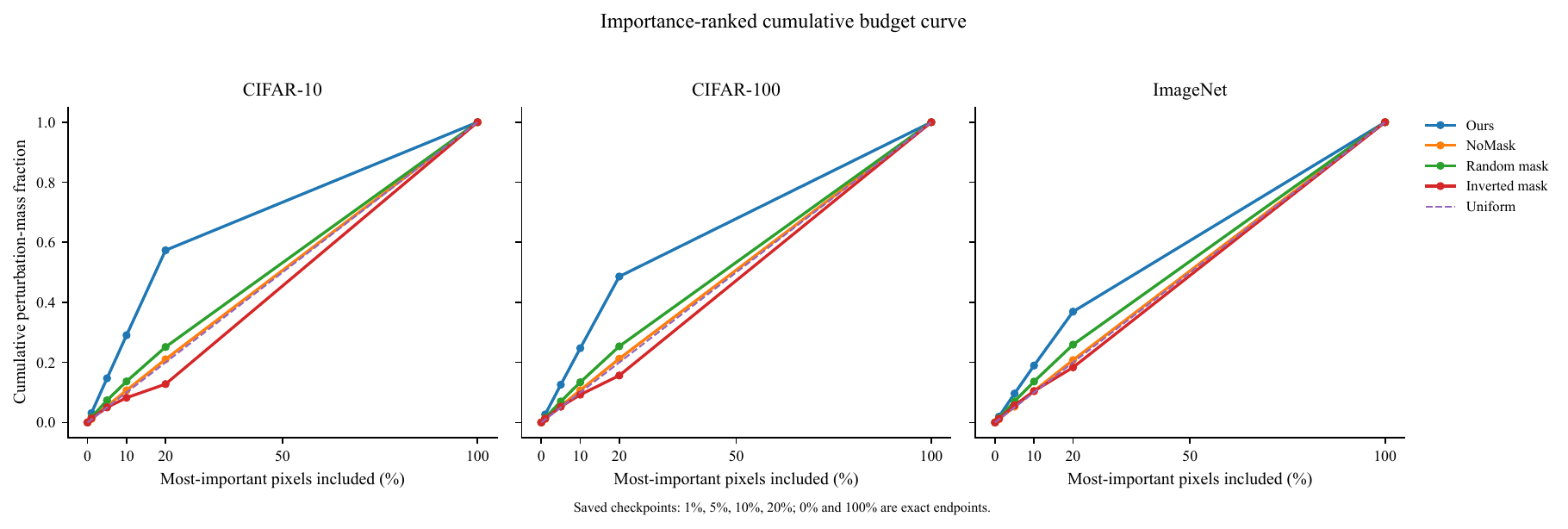}
    \caption{
    Cumulative perturbation mass after ordering locations from highest to
    lowest clean-gradient importance. The dashed diagonal corresponds to
    uniform allocation.
    }
    \label{fig:appendix_cumulative_mass}
\end{figure*}

%% file: Appendix/perceptual_and_norm_diagnostics.tex
\section{Perceptual and Norm Diagnostics}
\label{app:perceptual_diagnostics}

The proposed mechanism is designed to redistribute a fixed global
perturbation budget, not to optimize perceptual similarity. We therefore
report both global norm utilization and local/perceptual diagnostics.

\begin{table*}[t]
\centering
\caption{
Dataset-level macro-averaged perturbation diagnostics. Mean-$\ell_1$ and
$\ell_\infty$ are multiplied by 255. Lower LPIPS/DISTS and higher SSIM
indicate greater perceptual similarity.
}
\label{tab:perceptual_diagnostics}
\small
\begin{tabular}{llrrrrr}
\toprule
Dataset & Method
& Mean-$\ell_1{\times}255$
& $\ell_\infty{\times}255$
& LPIPS
& DISTS
& SSIM \\
\midrule
CIFAR-10
& APGD-CE  & 4.000 & 11.49 & 0.00223 & 0.06037 & 0.98126 \\
& APGD-DLR & 3.999 & 11.86 & 0.00255 & 0.06406 & 0.97999 \\
& PMA      & 4.000 & 10.81 & 0.00226 & 0.06066 & 0.98217 \\
& Ours     & 3.907 & 16.78 & 0.00296 & 0.07555 & 0.95456 \\
\midrule
CIFAR-100
& APGD-CE  & 4.000 & 11.45 & 0.00257 & 0.06383 & 0.97816 \\
& APGD-DLR & 3.997 & 12.27 & 0.00300 & 0.06826 & 0.97576 \\
& PMA      & 4.000 & 10.89 & 0.00266 & 0.06490 & 0.97871 \\
& Ours     & 3.959 & 16.18 & 0.00385 & 0.07855 & 0.95442 \\
\midrule
ImageNet
& APGD-CE  & 4.000 & 13.32 & 0.04236 & 0.08282 & 0.94746 \\
& APGD-DLR & 3.995 & 13.67 & 0.04245 & 0.08374 & 0.94562 \\
& PMA      & 4.000 & 11.54 & 0.04160 & 0.08208 & 0.94995 \\
& Ours     & 3.968 & 16.94 & 0.05060 & 0.10254 & 0.93714 \\
\bottomrule
\end{tabular}
\end{table*}

The realized global mean-$\ell_1$ magnitude of our method is not larger
than that of the matched baselines. However, its realized
$\ell_\infty$ magnitude is consistently larger, reflecting the intended
redistribution of perturbation toward selected regions.

The perceptual metrics similarly show that the increase in attack
effectiveness is not a free improvement in visual similarity. The proposed
method generally has larger LPIPS and DISTS and lower SSIM than the matched
baselines. These differences are consistent with stronger local
concentration while remaining within the same global mean-$\ell_1$ and
coordinate-cap constraints.

We therefore use these measurements as diagnostics of the perturbation
profile rather than as an additional performance claim.

%% file: Appendix/computational_cost.tex
\section{Computational Cost}
\label{app:compute}

Table~\ref{tab:computational_cost} reports model-evaluation counts and
dataset-level macro-averaged attack runtime.

\begin{table*}[t]
\centering
\caption{
Computational accounting for the main methods. Runtime is the recorded mean
attack runtime in seconds; it is intended for within-experiment comparison
rather than as hardware-independent benchmarking.
}
\label{tab:computational_cost}
\small
\begin{tabular}{llrrr}
\toprule
Dataset & Method
& Backward evals.
& Forward evals.
& Runtime (s) \\
\midrule
CIFAR-10
& APGD-CE  & 50 & 53  & 2.86 \\
& APGD-DLR & 50 & 53  & 2.88 \\
& PMA      & 50 & 102 & 4.08 \\
& Ours     & 51 & 103 & 4.06 \\
\midrule
CIFAR-100
& APGD-CE  & 50 & 53  & 2.42 \\
& APGD-DLR & 50 & 53  & 2.44 \\
& PMA      & 50 & 102 & 3.43 \\
& Ours     & 51 & 103 & 3.42 \\
\midrule
ImageNet
& APGD-CE  & 50 & 53  & 3.29 \\
& APGD-DLR & 50 & 53  & 3.29 \\
& PMA      & 50 & 102 & 4.73 \\
& Ours     & 51 & 103 & 4.71 \\
\bottomrule
\end{tabular}
\end{table*}

The additional cost of our method relative to PMA consists primarily of the
single clean-gradient computation used to construct the importance map.
Accordingly, the recorded runtime is very close to PMA.

Relative to the APGD variants, however, our implementation uses
substantially more forward evaluations. For this reason we describe the
method as approximately PMA-cost, rather than as universally
low-overhead.

The compute-matched $49+1$ control in
Appendix~\ref{app:mechanism_ablations} shows that the additional
clean-gradient backward evaluation does not explain the attack-success gain.

%% file: Appendix/statistical_analysis.tex
\section{Statistical Analysis}
\label{app:statistics}

All main attack comparisons are paired because methods within a
dataset--model configuration are evaluated on the same clean-correct
examples.

For each baseline $B$, we define the paired ASR effect
\begin{equation}
\Delta_{\mathrm{ASR}}
=
\mathrm{ASR}_{\mathrm{ours}}
-
\mathrm{ASR}_{B}.
\end{equation}

We report paired 95\% confidence intervals produced by the evaluation
pipeline. Differences in paired binary success outcomes are tested using
the exact McNemar test \cite{mcnemar1947note}. The family of main
comparisons is corrected using Holm's procedure
\cite{holm1979simple}.

\begin{table*}[t]
\centering
\caption{
Full paired main-result statistics. $\Delta$ is
$\mathrm{ASR}_{\mathrm{ours}}-\mathrm{ASR}_{\mathrm{baseline}}$ in
percentage points. $p_{\mathrm{Holm}}$ is the globally Holm-corrected exact
McNemar $p$-value.
}
\label{tab:full_pairwise_statistics}
\scriptsize
\begin{tabular}{lllrrl}
\toprule
Dataset & Model & Baseline
& $\Delta$
& 95\% CI
& $p_{\mathrm{Holm}}$ \\
\midrule
CIFAR-10 & WRN-94-16 & APGD-CE  & 9.34 & [8.76, 9.93] & $2.22\times10^{-262}$ \\
CIFAR-10 & WRN-94-16 & APGD-DLR & 7.70 & [7.15, 8.23] & $2.08\times10^{-208}$ \\
CIFAR-10 & WRN-94-16 & PMA      & 8.13 & [7.58, 8.70] & $4.09\times10^{-224}$ \\
CIFAR-10 & WRN-28-10 & APGD-CE  & 9.44 & [8.77, 10.11] & $3.86\times10^{-212}$ \\
CIFAR-10 & WRN-28-10 & APGD-DLR & 7.45 & [6.85, 8.05] & $3.91\times10^{-160}$ \\
CIFAR-10 & WRN-28-10 & PMA      & 7.79 & [7.19, 8.41] & $7.86\times10^{-171}$ \\
CIFAR-10 & XCiT-S12 & APGD-CE   & 7.44 & [6.91, 7.95] & $1.39\times10^{-196}$ \\
CIFAR-10 & XCiT-S12 & APGD-DLR  & 5.49 & [5.02, 6.00] & $8.70\times10^{-125}$ \\
CIFAR-10 & XCiT-S12 & PMA       & 5.49 & [5.01, 5.96] & $2.08\times10^{-127}$ \\
\midrule
CIFAR-100 & WRN-28-10 & APGD-CE  & 12.87 & [12.07, 13.61] & $1.79\times10^{-280}$ \\
CIFAR-100 & WRN-28-10 & APGD-DLR & 6.96 & [6.31, 7.57] & $1.04\times10^{-122}$ \\
CIFAR-100 & WRN-28-10 & PMA      & 7.82 & [7.18, 8.48] & $2.54\times10^{-153}$ \\
CIFAR-100 & XCiT-S12 & APGD-CE   & 8.26 & [7.62, 8.90] & $2.92\times10^{-159}$ \\
CIFAR-100 & XCiT-S12 & APGD-DLR  & 2.73 & [2.15, 3.34] & $1.46\times10^{-19}$ \\
CIFAR-100 & XCiT-S12 & PMA       & 2.52 & [2.00, 3.06] & $1.47\times10^{-19}$ \\
CIFAR-100 & WRN-70-16 & APGD-CE  & 12.38 & [11.62, 13.13] & $7.46\times10^{-277}$ \\
CIFAR-100 & WRN-70-16 & APGD-DLR & 7.15 & [6.57, 7.78] & $2.57\times10^{-137}$ \\
CIFAR-100 & WRN-70-16 & PMA      & 7.71 & [7.09, 8.34] & $3.59\times10^{-154}$ \\
\midrule
ImageNet & ConvNeXt-L & APGD-CE  & 18.00 & [15.60, 20.40] & $6.57\times10^{-52}$ \\
ImageNet & ConvNeXt-L & APGD-DLR & 18.20 & [15.90, 20.80] & $6.85\times10^{-49}$ \\
ImageNet & ConvNeXt-L & PMA      & 15.90 & [13.70, 18.20] & $9.98\times10^{-46}$ \\
ImageNet & Peng et al. & APGD-CE  & 19.80 & [17.40, 22.20] & $6.97\times10^{-59}$ \\
ImageNet & Peng et al. & APGD-DLR & 20.10 & [17.60, 22.60] & $4.13\times10^{-58}$ \\
ImageNet & Peng et al. & PMA      & 17.70 & [15.30, 20.00] & $1.25\times10^{-52}$ \\
ImageNet & ViT-B-ConvStem & APGD-CE  & 13.00 & [11.00, 15.00] & $1.18\times10^{-38}$ \\
ImageNet & ViT-B-ConvStem & APGD-DLR & 13.20 & [11.10, 15.50] & $1.59\times10^{-34}$ \\
ImageNet & ViT-B-ConvStem & PMA      & 9.90 & [8.10, 12.00] & $3.31\times10^{-25}$ \\
ImageNet & Swin-L & APGD-CE  & 10.20 & [8.40, 12.20] & $6.21\times10^{-29}$ \\
ImageNet & Swin-L & APGD-DLR & 8.70 & [6.80, 10.70] & $3.26\times10^{-19}$ \\
ImageNet & Swin-L & PMA      & 7.80 & [6.00, 9.60] & $1.08\times10^{-18}$ \\
\bottomrule
\end{tabular}
\end{table*}

All 30 paired ASR differences are positive, all reported confidence
intervals exclude zero, and all comparisons remain significant after the
global Holm correction. The largest corrected $p$-value is approximately
$1.08\times10^{-18}$.

Because the evaluation populations are large, we emphasize the effect sizes
and confidence intervals rather than the magnitude of extremely small
$p$-values.

%% file: Appendix/descriptive_cross_regime_results.tex
\section{Descriptive Cross-Regime Results}
\label{app:cross_regime}

This section preserves the conventional-versus-capacity-limited reference
experiment from the earlier study design. These results are intentionally
separated from the main empirical evidence.

The two regimes use different global-budget parameterizations and therefore
do \emph{not} constitute a controlled intervention in which only the local
capacity changes.

\subsection{Budget Parameterizations}

For conventional $\ell_1$-APGD and FMN-$\ell_1$, the total
$\ell_1$ radius is
\begin{equation}
\epsilon_1 = 12
\end{equation}
on CIFAR and
\begin{equation}
\epsilon_1 = 60
\end{equation}
on ImageNet.

The proposed conventional variant instead uses the mean-$\ell_1$
parameterization
\begin{equation}
\frac{\|\delta\|_1}{D}\leq\frac{4}{255}.
\end{equation}

Under the capacity-limited setting, all reported methods use
\begin{equation}
\frac{\|\delta\|_1}{D}\leq\frac{4}{255}
\end{equation}
together with the heterogeneous local-capacity envelope
\begin{equation}
\mu_E=\frac{12}{255},
\qquad
\sigma_E=\frac{2}{255}.
\end{equation}

Thus, both the presence of local capacities and the global-budget
parameterization differ across portions of the conventional comparison.

\subsection{Reference Results}

\begin{table*}[t]
\centering
\caption{
Descriptive results under separately parameterized conventional and
capacity-limited $\ell_1$ threat settings. Because the global budget
parameterizations differ, changes between the two column groups must not be
interpreted as the isolated causal effect of adding local capacities.
}
\label{tab:descriptive_cross_regime}
\small
\begin{tabular}{llrccc|ccc}
\toprule
& & &
\multicolumn{3}{c}{Conventional $\ell_1$ ASR (\%)}
&
\multicolumn{3}{c}{Capacity-limited ASR (\%)} \\
Dataset & Model & $N$
& $\ell_1$-APGD & FMN-$\ell_1$ & Ours
& Adapted $\ell_1$-APGD & Adapted FMN-$\ell_1$ & Ours \\
\midrule
CIFAR-10
& WRN-94-16
& 9,368
& 84.03 & 81.18 & 18.01
& 32.55 & 22.88 & 17.98 \\
CIFAR-100
& WRN-70-16
& 7,522
& 88.86 & 82.82 & 32.03
& 49.19 & 43.23 & 32.53 \\
ImageNet
& Swin-L
& 1,000
& 44.50 & 33.90 & 39.50
& 69.00 & 16.50 & 36.00 \\
\bottomrule
\end{tabular}
\end{table*}

The table documents substantial method-dependent differences across the two
separately defined regimes. For example, $\ell_1$-APGD decreases on the
two CIFAR reference configurations but increases on ImageNet, whereas
FMN-$\ell_1$ decreases on all three.

These observations are useful descriptively, but they do not establish that
local capacities alone caused the changes. We therefore do not use this
experiment to support the main claim of the paper.

A second observation is important for the scope of our contribution:
adapted specialized $\ell_1$-APGD is stronger than our reference optimizer
on all three capacity-limited configurations in this table. The proposed
method should consequently be interpreted as an allocation mechanism that
improves the matched APGD/PMA-style optimizers studied in the main
experiment, rather than as a universally strongest $\ell_1$ attack.

%% file: Appendix/constraint_validity_checks.tex
\section{Constraint-Validity Checks}
\label{app:constraint_validity}

We verify the capacity-limited threat constraints for every method in the
main ten-configuration evaluation.

For every generated perturbation we check

\begin{align}
\frac{\|\delta\|_1}{D}
&\leq
\frac{4}{255},
\\
|\delta_i|
&\leq
E_i
\qquad \forall i,
\\
0
&\leq
x_i+\delta_i
\leq
1
\qquad \forall i.
\end{align}

We additionally verify that matched methods use the intended shared
capacity and initialization assets through stored digests.

\begin{table*}[t]
\centering
\caption{
Constraint-validity summary across the ten main model--dataset
configurations. Numerical residuals are maximum observed floating-point
excesses above the corresponding constraints.
}
\label{tab:constraint_validity}
\small
\begin{tabular}{lrrrrr}
\toprule
Method
& Violations
& Max mean-$\ell_1$ excess
& Max cap excess
& Image-low excess
& Image-high excess \\
\midrule
APGD-CE
& 0
& $3.07\times10^{-8}$
& $2.98\times10^{-8}$
& 0 & 0 \\
APGD-DLR
& 0
& $2.89\times10^{-8}$
& $2.98\times10^{-8}$
& 0 & 0 \\
PMA
& 0
& $5.68\times10^{-8}$
& $2.98\times10^{-8}$
& 0 & 0 \\
Ours
& 0
& $3.26\times10^{-8}$
& $2.98\times10^{-8}$
& 0 & 0 \\
\bottomrule
\end{tabular}
\end{table*}

No substantive constraint violations are observed. The nonzero residuals
are on the order of $10^{-8}$ and are attributable to floating-point
precision.

Across the main comparisons, the recorded capacity-digest mismatch count
and initialization-digest mismatch count are both zero. The ablation
projection preflight checks additionally pass the $2\times10^{-6}$
numerical tolerance.

These diagnostics support the interpretation that differences in attack
behavior arise within the intended common feasible region rather than from
constraint violations or mismatched capacity realizations.

%% file: Appendix/additional_qualitative_examples.tex
\section{Additional Qualitative Examples}
\label{app:qualitative_examples}

Qualitative visualizations complement the quantitative allocation
measurements by showing how different mechanism variants distribute
perturbation under the same threat constraints.

For each displayed example, we visualize:

\begin{enumerate}
    \item the clean input;
    \item the fixed clean-gradient importance map;
    \item the perturbation magnitude produced without allocation guidance;
    \item the perturbation produced using a randomized prior;
    \item the perturbation produced using an inverted prior;
    \item the perturbation produced using a smoothed importance map; and
    \item the perturbation produced by the full centered importance-guided
    method.
\end{enumerate}

Perturbation visualizations are normalized consistently within an example
for display purposes. This visualization normalization does not alter the
actual perturbations used for evaluation.

The representative example used in the main qualitative analysis is
selected using the predefined median-LPIPS criterion rather than by manually
choosing the most visually favorable case. Additional examples should use
the same deterministic selection principle.



The qualitative examples should not be interpreted as independent evidence
of attack superiority. Their purpose is to make the allocation behavior
measured by Top-10 mass, cumulative importance-ranked mass, effective
support, and spatial Gini visually interpretable.